\documentclass{article}

\usepackage{microtype}
\usepackage{graphicx}
\usepackage{subcaption}
\usepackage{booktabs} 
\usepackage{hyperref}

\usepackage[preprint]{icml2026}

\usepackage{amsmath}
\usepackage{amssymb}
\usepackage{mathtools}
\usepackage{amsthm}
\usepackage{csquotes}
\usepackage{multirow}
\usepackage{tabularx}

\usepackage[most]{tcolorbox}
\usepackage{xcolor}
\usepackage{listings}
\usepackage{hyperref}
\hypersetup{hidelinks}
\usepackage[capitalize,noabbrev,nameinlink]{cleveref}

\crefname{lstlisting}{listing}{listings}
\Crefname{lstlisting}{Listing}{Listings}

\definecolor{lightgray}{gray}{0.95}

\newtcolorbox{hypobox}[1]{%
  colback=lightgray,
  colframe=black!15,
  boxrule=0.5pt,
  arc=2mm,
  left=3mm, right=3mm, top=2mm, bottom=2mm,
  fonttitle=\bfseries,
  coltitle=black,     
  title={#1}
}

\definecolor{conclusionblue}{HTML}{282561}

\newtcolorbox{conclusionbox}[1]{%
  colback=conclusionblue!18,       
  colframe=conclusionblue,
  boxrule=0.7pt,
  arc=2mm,
  left=3mm, right=3mm, top=2mm, bottom=2mm,
  fonttitle=\bfseries,
  coltitle=black,
  colbacktitle=conclusionblue!25,  
  title={#1}
}
\theoremstyle{plain}

\theoremstyle{definition}

\theoremstyle{remark}

\usepackage[textsize=tiny]{todonotes}

\icmltitlerunning{Semantic Content Determines Algorithmic Performance}

\begin{document}

\twocolumn[
  \icmltitle{Semantic Content Determines Algorithmic Performance}



  \icmlsetsymbol{equal}{*}

  \begin{icmlauthorlist}
    \icmlauthor{Martiño Ríos-García}{yyy}
    \icmlauthor{Nawaf Alampara}{yyy}
    \icmlauthor{Kevin Maik Jablonka}{yyy,sch,comp,zzzz}
  \end{icmlauthorlist}

  \icmlaffiliation{yyy}{Laboratory of Organic and Macromolecular Chemistry (IOMC), Friedrich Schiller University Jena, Humboldtstrasse 10, 07743 Jena, Germany}
  \icmlaffiliation{comp}{HIPOLE Jena (Helmholtz Institute for Polymers in Energy Applications Jena), Lessingstrasse 12-14, 07743 Jena, Germany}
  \icmlaffiliation{sch}{Center for Energy and Environmental Chemistry Jena (CEEC Jena), Friedrich Schiller University Jena, Philosophenweg 7a, 07743 Jena, Germany}
  \icmlaffiliation{zzzz}{Jena Center for Soft Matter (JCSM), Friedrich Schiller University Jena, Philosophenweg 7, 07743 Jena, Germany}

  \icmlcorrespondingauthor{Kevin Maik Jablonka}{mail@kjablonka.com}

  \icmlkeywords{ICML, Machine Learning, Large Language Models, Evaluation, Benchmarks, Robustness, Semantics, Counting}

  \vskip 0.3in
]



\printAffiliationsAndNotice{}  

\begin{abstract}
    Counting should not depend on what is being counted; more generally, any algorithm's behavior should be invariant to the semantic content of its arguments. We introduce WhatCounts to test this property in isolation. Unlike prior work that conflates semantic sensitivity with reasoning complexity or prompt variation, WhatCounts is atomic: count items in an unambiguous, delimited list with no duplicates, distractors, or reasoning steps for different semantic types. Frontier LLMs show over 40~\% accuracy variation depending solely on what is being counted---cities versus chemicals, names versus symbols. Controlled ablations rule out confounds. The gap is semantic, and it shifts unpredictably with small amounts of unrelated fine-tuning. LLMs do not implement algorithms; they approximate them, and the approximation is argument-dependent. As we show with an agentic example, this has implications beyond counting: any LLM \enquote{function} may carry hidden dependencies on the meaning of its inputs.
\end{abstract}

\section{Introduction}

An algorithm's behavior should depend on the structure of its input, not the meaning. Counting ten items should work identically whether they are cities or chemicals, names or emojis. This invariance is what makes an algorithm an algorithm \cite{ouellette2023counting, zhou2024what}.

\begin{figure*}[ht]
  \centering
  \includegraphics[width=\textwidth]{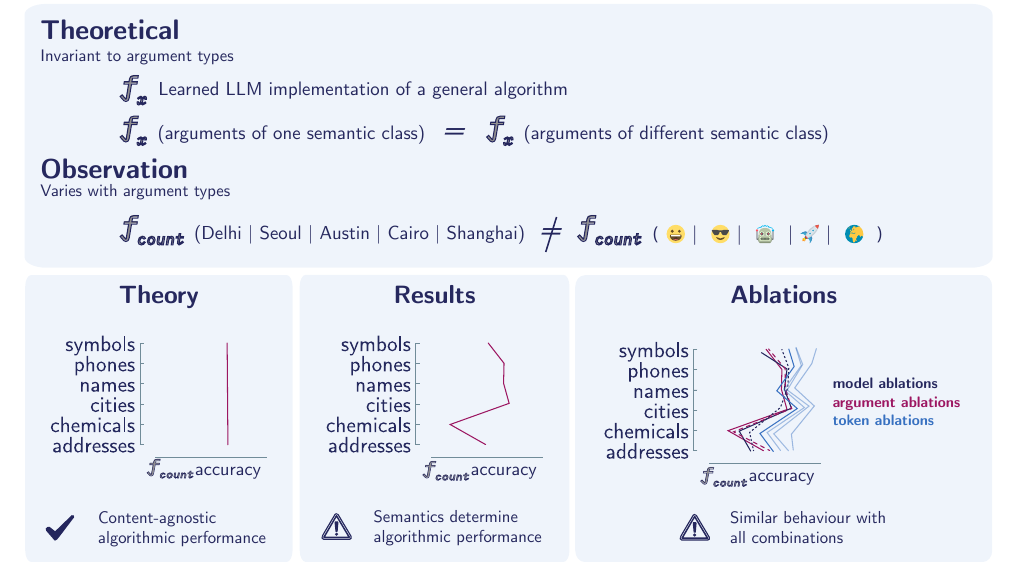}
  \caption{\textbf{Overview of WhatCounts}. The top panel contrasts the theoretical ideal with the observed behavior of LLMs. In theory, an algorithm (e.g., counting) should be invariant to the meaning of its arguments, performing identically for lists of cities or emojis. Results indicate that accuracy varies strongly across semantic categories. We further run controlled ablations (e.g., token shuffling, token-count controls, and explicit separator specification), none of which eliminate the semantic class dependence, effectively demonstrating the inherent semantic fragility.}
  \label{fig:overview}
\end{figure*}

LLMs appear to implement many such algorithmic operators; for example, counting, sorting, searching, and extracting \cite{schick2023toolformer, herbold2025sortbench0, suzgun2022challenging, srivastave2023beyond}. Some works have even reported that we are now in the age of \enquote{prompt programming} \cite{liang2025prompts, reynolds2021prompt}.
But do they actually implement them, or merely approximate them? If an LLM implements a counting procedure, its errors should reflect general limitations, such as context limits, not systematic dependence on the meaning of what is being counted \cite{mirzadeh2024gsm0symbolic0, sclar2024quantifying}. If errors vary with semantic content, the model is not executing an algorithm; it is rather pattern-matching one \cite{weiss2021thinking}.
This matters as LLMs are increasingly deployed as components in production systems, where pipelines chain prompt-based subroutines, agents gate tool calls, assuming that low-level operators behave predictably across domains. In such settings, primitive algorithmic tasks serve as building blocks, and content-conditioned failures propagate into system-level brittleness.

We directly test the sensitivity of LLMs on the semantic class of their arguments. WhatCounts is a minimal benchmark. It counts items in an unambiguous, delimited list containing only valid, non-duplicate single semantic classes. Prior work has established that variations of the prompt template can lead to surprising changes in the performance---but tacitly assumed that the semantic class of the arguments or template variables does not. WhatCounts isolates this single question: does the \emph{semantic content} of the arguments affect the procedure's success? 

We show that it does. We surprisingly show that frontier LLMs show over 40\% accuracy variation depending solely on the semantic class under identical formatting and identical list lengths. We further demonstrate that these variations persist even in agentic environments with a Python execution tool available. 

We rule out surface explanations. Models correctly segment and identify semantic classes they cannot count: when prompted to wrap items in XML tags, they succeed with high reliability. Providing explicit structure does not help: even with tags around each item, the semantic gaps persist. Controlling for token count rather than item count reproduces the same patterns. When reducing semantics by shuffling tokens within items in the list---corrupting meaning while preserving statistics---performance drops, suggesting that semantics influence algorithmic performance.

We further show that these biases are unstable. Comparing models fine-tuned under identical regimes but on different datasets, we find that semantic class-specific failure patterns shift unpredictably. Training on data unrelated to counting nonetheless reshapes which semantic classes succeed and which fail, showing that WhatCounts measures an important fragility mode that other evaluations ignored so far.

These findings have practical consequences. Counting is a primitive in (LLM) pipelines \cite{gao2023pal}. If counting behaves as a content-conditioned heuristic rather than an invariant operator, downstream systems inherit this dependence. In this way, a seemingly trivial subroutine becomes a systematic source of unreliability.

\paragraph{Contributions.}
\begin{enumerate}
    \item \textbf{WhatCounts}, an atomic benchmark isolating semantic invariance in counting \footnote{Code available in \url{https://github.com/lamalab-org/whatcounts}}---free from the confounds of tokenization, ambiguity, and multi-step reasoning.
    
    \item \textbf{Semantic content determines algorithmic performance.} Counting accuracy varies by over 40\% across semantic class types under identical conditions. Controlled ablations rule out tokenization, item recognition, and sequence length.
    
    \item \textbf{Unstable failure modes.} Semantic class biases shift unpredictably with fine-tuning on unrelated data, confirming the semantic gap is caused by the model weights and not the architecture.
    
    \item \textbf{Downstream risk.} Semantic fragility in a basic operator undermines production reliability as we show with an agentic use case.
\end{enumerate}


\section{Related Work}

\subsection{Algorithmic Operators in Transformers}

A central theoretical question is what it means, mechanistically, for a Transformer to implement an algorithmic operator, and which architectural resources determine when such implementations are possible.

\citet{weiss2021thinking} introduced RASP, a programming language built around primitives that can be implemented with Transformers. 
\citet{zhou2024what} showed how algorithms that can be implemented in a simple program in the RASP-L programming language can be more easily learned by a Transformer than those that require a long program.

Complementary work provides limitations and capacity requirements.
For counting in particular, \citet{yehudai2024transformers} studied simple token-frequency counting and showed a sharp capacity threshold: counting can be implemented when the model dimension scales linearly with context length, with theory and experiments indicating failure below this regime.
Other works connected Transformer variants to circuit- and logic-based characterizations, showing that seemingly small choices---e.g., precision assumptions, circuit uniformity, or allowing intermediate decoding (CoT)---can change which functions are efficiently computable \cite{merrill2024expressive, chiang2023tighter, chiang2024transformers}.

In contrast to these works, which examined the architectural capabilities from theoretical and empirical perspectives, our work directly evaluates whether a learned approximation of an algorithmic operator is invariant to the semantic class of its arguments.

\begin{figure*}[tbp]
  \centering
  \includegraphics[width=\textwidth]{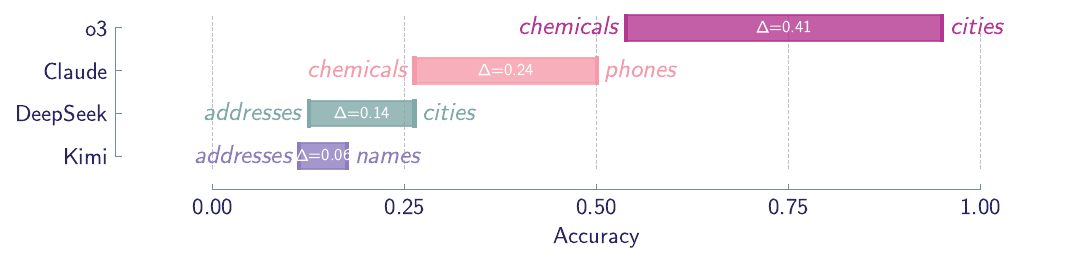}
  \caption{\textbf{The semantic gap varies significantly across models, with top performers showing the highest variance between minimum and maximum semantic class scores}. The horizontal bars represent the semantic gap (the smaller, the better), with the classes at the ends of the bars representing the minimum and maximum scores for each model. \Cref{fig:general_app} shows the detailed results for each semantic class. We find that the better-performing models are also more susceptible to changes in semantic class.}
  \label{fig:fig-1}
\end{figure*}

\subsection{Counting}

Given the foundational nature of counting, several works have evaluated LLMs' counting-related abilities across diverse settings and benchmarks \cite{xu2025llm, chang2025language, fu2024large, srivastava2022bigbench}. 
A widely discussed example is character counting (e.g., counting the number of r's in \enquote{strawberry}), which is strongly influenced by tokenization and subword segmentation \cite{cosma2025strawberry, zhang2024counting}. 
Motivated by these findings, we focus on item-level enumeration in lists.

Most closely related to our work, \citet{ball2024canwe} studied deterministic tasks such as list counting and selection operators (e.g., maximum, median, sorting) and showed that performance varies not only with prompt phrasing and list length, but also with the list composition and object frequency, highlighting a \enquote{language-as-fixed-effect} concern in capability measurement, as prompt fragility is a largely-ignored source of error.

Our work aims to isolate a completely different axis: \textit{semantic invariance}. We remove common sources of ambiguity by using unambiguous, delimited lists with no duplicates and no distractors, and perform targeted controls to rule out tokenization and other possible explanations. While \citet{ball2024canwe} demonstrated that prompt phrasing and list composition affect performance, we hold these factors constant to isolate whether semantically equivalent input lists that differ only in what is being counted yield systematically different behaviors. 
This design allows us to measure a robust semantic class dependence in counting accuracy under conditions where, from an operator viewpoint, semantics should be irrelevant. We also provide WhatCounts as a benchmark implementation that can be used efficiently to estimate semantic invariance. In addition, we demonstrate that the effects we measure with WhatCounts have repercussions in agentic use cases. 

\subsection{LLM Fragility}

LLM fragility under minor, meaning- and complexity-preserving input changes has been widely documented \cite{sclar2024quantifying, errica2025what}. 
Robustness can also degrade under realistic noise, such as typographical errors \cite{haller2025llm, liu2025evaluating}, or under lexical substitutions, such as the use of synonyms \cite{Alahmari2025large, lin2024llm}.

In mathematical reasoning, GSM-Symbolic \cite{mirzadeh2024gsm0symbolic0} built controlled variants of GSM8K-style problems \cite{cobbe2021training} using symbolic templates, and showed substantial variance across different instances of the same task. Similarly, Alice in Wonderland \cite{nezhurina2024alice} demonstrated severe errors on a short commonsense reasoning template and analyzed some common mitigation strategies.

Outside of mathematics, \citet{ullman2023large} showed that LLMs can fail on theory-of-mind (ToM) tasks, where the reasoning of the models changed when changing the problem formulation. 

Overall, prior work leaves little doubt that LLM behavior can change substantially under small, seemingly benign transformations. 
However, in most studied settings, these variations are inseparable from other effects: domain knowledge (e.g., mathematics or commonsense), multi-step reasoning, ambiguity resolution, or distractor handling. In addition, they focus on specific variables in the task templates, assuming the semantic class of the arguments filled into the templates does not matter.
As a result, it remains difficult to determine whether the observed sensitivity reflects brittleness of the underlying operator the model is meant to implement, or failures in the surrounding skills needed to solve the task instance. 
WhatCounts addresses this gap by providing an atomic probe that separates all effects other than semantic invariance, and probing how much such semantics can impact agentic tasks.

\section{Results} \label{sec:main-results}

We chose six common semantic classes for evaluation: addresses, chemicals, cities, full names of people, phone numbers, and emojis. A detailed description of the procedure for obtaining each is provided in \Cref{sec:obtain-entity}. For each experiment, we select different count ranges to avoid the tendency of models to respond with rounded results (see \Cref{sec:round-answer} for more detail on this). 
We considered 20 different counts per task (one specific entity class and range) to account for the statistical variance. The items in the list are randomly sampled from the databases, and separated by the pipe (\textbar, see more discussion on the separator in \Cref{sec:sep-changes}). All items in the list count, meaning we did not consider distractors, ensuring no duplicates are present (prompt available in \Cref{lst:counting-prompt}). With this, we ensure we measure only the models' ability to count. This enables us to perform a clean measurement of the semantic gap.

We evaluated four frontier models---two open-source and two proprietary: \texttt{claude-sonnet-4-20250514} from Anthropic, \texttt{deepseek-v3}, \texttt{kimi-k2-instruct-0905} from Moonshot AI, and \texttt{o3-2025-04-16} from OpenAI. We run the models with the recommended settings unless otherwise mentioned, as detailed in \Cref{sec:model-settings}. 

As a measure of how sensitive the models are to semantic changes in the lists, we introduce the \textit{Semantic Gap}, $\Delta_{\mathrm{sem}}(m)$, which is defined as follows:

\begin{equation} \label{eq:semantic-gap}
\Delta_{\mathrm{sem}}(m)
=
\max_{e \in \mathcal{E}} \, \mathrm{Acc}(m,e)
-
\min_{e \in \mathcal{E}} \, \mathrm{Acc}(m,e).
\end{equation}

\noindent where the $\mathcal{E}$ is the set of all semantic classes considered in the benchmark, and $\mathrm{Acc}(m,e)\in[0,1]$ is the accuracy achieved by the model $m$ for the evaluation of the class $e$. We considered correctness as binary.

\Cref{fig:fig-1} shows the results of the models in the WhatCounts evaluation (the detailed results per semantic class can be found in \Cref{fig:general_app}). 

Our results show that the better the model performs on average for the counting task, the bigger the semantic gap. 
This does not align with common sense, and it is surprising, as one would expect stronger models to be less prone to any form of fragility \cite{mirzadeh2024gsm0symbolic0, nezhurina2024alice}.

This semantic gap can have direct implications for evaluations, as the models' scores can be greatly affected by the semantics used in the tasks. Future benchmarks should account for this effect to properly assess performance and robustness.

\section{Ablation Experiments}

We distinguish structural properties---list length, token count, formatting---from semantic category: the ontological class being processed. An algorithm, by definition, should depend on the former but not the latter. We designed a set of ablation experiments to isolate the structural properties from the semantic class to evaluate what leads to the algorithm failure.

\begin{figure}[hb]
  \centering
  \includegraphics[width=\columnwidth]{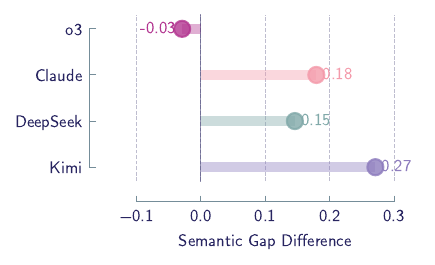}
  \caption{\textbf{Token-controlled lists ablation reveals that semantic gaps are wider when fixing tokens compared to fixing classes counts.} The bars represent the semantic gap difference with respect to the result of the list count being fixed. Negative results indicate that the semantic gap was reduced when fixing the token count instead of item count. We observe that the semantic gap generally increases when fixing the token counts across classes in the list, with the notable exception of o3.}
  \label{fig:fig-2}
\end{figure}

\begin{figure*}[!ht]
  \centering
  \includegraphics[width=\textwidth]{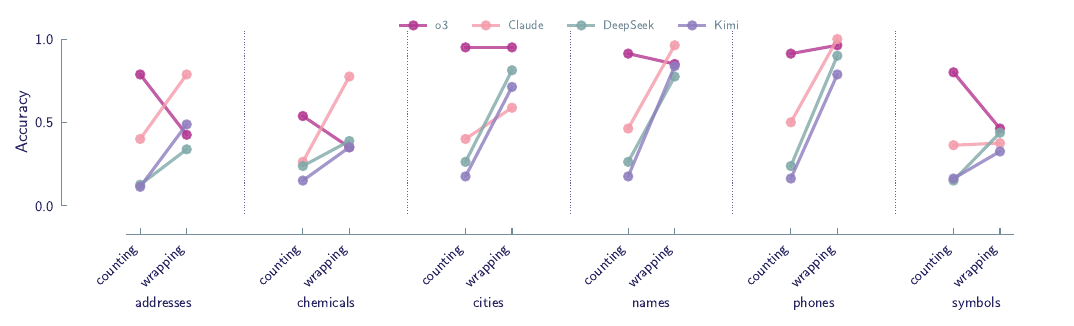}
  \caption{\textbf{Most models demonstrate superior performance in identifying and wrapping items of the different classes compared to basic counting.} The results show that, models, except for \texttt{o3}, understand semantic classes better than they can count them.}
  \label{fig:wrapping}
\end{figure*}

\vspace{\baselineskip}
\vspace{\baselineskip}
\vspace{\baselineskip}
\vspace{\baselineskip}
\vspace{\baselineskip}
\vspace{\baselineskip}

\textbf{Token-count Dependence} 
\begin{hypobox}{Hypothesis 1}
    The difference in token counts between semantical classes can influence the results.
\end{hypobox}

To investigate this, we ensured that the prompts for the different classes contain the same number of tokens.

In \Cref{fig:fig-2} (detailed results in \Cref{fig:token_app}) we show how the semantic gap changes in the token count-controlled setting compared to the item count-controlled setting we discussed in \Cref{fig:fig-1}.
The figure shows that when equating the number of tokens across classes, the semantic gap increases for all models (except for o3, for which the difference is minimal), indicating that the semantic variability cannot be explained by differences in token counts across semantic classes. This is, that by fixing the tokens in the lists, all models still exhibit semantic gaps exceeding 25\%.

\begin{conclusionbox}{Token-wise Conclusion}
    Token-count differences between semantic classes cannot explain the semantic gaps.
\end{conclusionbox}

\begin{figure*}[!ht]
  \centering
  \includegraphics[width=\textwidth]{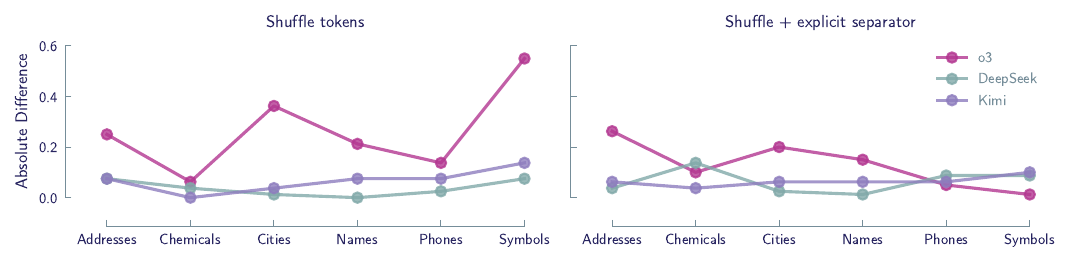}
  \caption{\textbf{The semantics that the tokens carry have some effect as there are substantial differences with respect to the non-shuffled case.} Absolute difference of shuffling the tokens, and not doing it. Note that Claude is not present in this figure because we cannot run the experiment without access to the tokenizer.}
  \label{fig:shuffling}
\end{figure*}

\begin{table}[h]
    \centering
    \caption{\textbf{Semantic gap when separator is explicitly mentioned, and the difference to when it is not mentioned.} We observe that there are no differences between mentioning or not mentioning the separator in the prompt. This shows that the specification of the separator is not the reason for the variability observed for the different classes.}
    \begin{tabularx}{\columnwidth}{l>{\centering\arraybackslash}X>{\centering\arraybackslash}X}
\toprule
Model & $\Delta_{\text{sem}}$ with Explicit Separator & Difference in $\Delta_{\text{sem}}$ \\
\midrule
o3 & 0.400 & -0.012 \\
Claude & 0.237 & +0.000 \\
DeepSeek & 0.087 & -0.050 \\
Kimi & 0.125 & +0.062 \\
\bottomrule
\end{tabularx}

    \label{tab:explicit_sep}
\end{table}

\textbf{Separator Dependence} 
\begin{hypobox}{Hypothesis 2}
    The models may be unable to recognize the structure of the list for some of the semantic classes.
\end{hypobox}

In the previous experiments, the separator between the items was not explicitly told to the model. 
This might make it difficult for the models to separate the different items in the list. To exclude this possible cofounder, we ask the models to count, explicitly mentioning the separator in the prompt (see \Cref{lst:explicit-prompt} for the instruction used).

\Cref{tab:explicit_sep} shows (detailed results in \Cref{fig:explicit_app}) the semantic gap we observe when the separator is explicitly mentioned. The values for the semantic gap are very similar to those without the separator being mentioned, indicating that the separator is not the problem.

\begin{conclusionbox}{Separator Dependence}
    The semantic gap is not produced because of not recognizing the list structure.
\end{conclusionbox}

\textbf{Isolating Identifyability} 

\begin{hypobox}{Hypothesis 3}
    Semantic classes differ in their ability to be identified by a model.
\end{hypobox}

Counting is not only about aggregating items, but also about identifying them first \cite{yehudai2024transformers}. 
A possible reason for the variation in performance across the different classes might be that the models have difficulty understanding the items' boundaries and, thus, identifying them.
To test whether the models can identify the items in the list, we ask them to return the input list, with each item wrapped in XML tags.

\Cref{fig:wrapping} (refer to \Cref{fig:wrap_app} for the elaborate results) shows the results when the models are instructed to wrap the items of the list compared to the counting task. We observe that models are substantially better at identifying the items that are counted, suggesting the problem lies in the aggregation of the counts.
However, the variance between semantic classes persists across all models, so while they do a better job at identification than in aggregation, the semantic dependence still persists.

\begin{conclusionbox}{Counting with Wrapped Items}
     The semantic gap for identification is lower than for aggregation.
\end{conclusionbox}

\textbf{Isolating Aggregation} 
\begin{hypobox}{Hypothesis 4}
    Giving the list with each item wrapped in XML tags homogenizes the inputs across semantic classes, such that the semantic gap will disappear.
\end{hypobox}
To isolate the aggregation performance, we modified the list of items, wrapping each item in XML tags to help the models better identify the individual items of the list.

\begin{table}[h]
    \centering
    \caption{\textbf{Semantic gap when the items in the list are wrapped between XML tags.} We see no major differences when wrapping the items in the list using XML tags with respect to the counting task in which the list is not wrapped, indicating again that identifiability of items is not the reason for the semantic variability between classes.}
    \begin{tabularx}{\columnwidth}{l>{\centering\arraybackslash}X>{\centering\arraybackslash}X}
\toprule
Model & $\Delta_{\text{sem}}$ with XML Tags & Difference in $\Delta_{\text{sem}}$ \\
\midrule
o3 & 0.412 & +0.000 \\
Claude & 0.250 & +0.013 \\
DeepSeek & 0.025 & -0.112 \\
Kimi & 0.075 & +0.012 \\
\bottomrule
\end{tabularx}

    \label{tab:xml_ablation}
\end{table}

\Cref{tab:xml_ablation} (see \Cref{fig:xml-app}) details the results of wrapping the list items for the different semantic classes between XML tags. 
The semantic gap of top models seems not to depend on whether items are wrapped in XML tags (see \Cref{fig:fig-1}). Only DeepSeek seems to show a change in the semantic gap, but it does not vanish completely. 

\begin{conclusionbox}{Counting with Wrapped Items}
    The semantic gap does not change when wrapping the input items between XML tags. Thus, even the aggregation of clearly identified items shows a semantic gap.
\end{conclusionbox}

\textbf{Token-distribution Dependence} 
\begin{hypobox}{Hypothesis 5}
    If we shuffle the tokens of each item in the list, the algorithm's performance should be the same if it is invariant to the semantics of its arguments.
\end{hypobox}

If the LLM function is invariant to changes in the semantic classes when tokens are shuffled, the performance should not change unless there is some effective dependence on the semantic classes. For that, we simply run the same experiment but shuffle the tokens of each item in the list (prompt detailed in \Cref{lst:counting-prompt}). Additionally, we also run the analogous experiment,  additionally mentioning the explicit separator (prompt shown in \Cref{lst:explicit-prompt}).

\Cref{fig:shuffling} (see \Cref{sec:shuffling-app} for more details) represents the performance in the shuffling experiments. We see that there are substantial differences, especially for the stronger model.

\begin{conclusionbox}{Token-distribution Dependence}
    The variations in performance depend on the semantics, as when shuffling the tokens, despite the problem being invariant to this change, the LLM's output is different.
\end{conclusionbox}

\textbf{Reasoning Dependence} 
\begin{hypobox}{Hypothesis 6}
    Increasing reasoning effort reduces the semantic gap as the model can leverage more involved counting \enquote{algorithms}.
\end{hypobox}

To investigate this, we run the \texttt{o3} model with the different \enquote{reasoning efforts} enabled. 

\begin{figure}[!hb]
  \centering
  \includegraphics[width=\columnwidth]{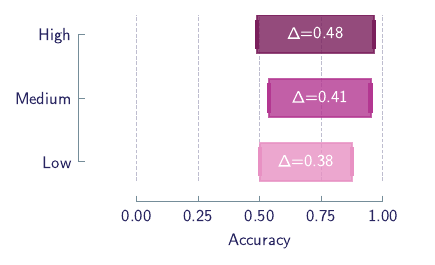}
  \caption{\textbf{Reasoning effort affects the semantic gap with minimal impact on accuracy.} The bars expand between the minimum and the maximum accuracy of the accuracy value for the different semantic classes, representing the semantic gap annotated inside. The results show that reasoning effort does not affect accuracy, but it seems to affect the semantic gap, which increases with greater effort.}
  \label{fig:fig-reasoning}
\end{figure}

\Cref{fig:fig-reasoning} shows the results of varying the reasoning effort of the \texttt{o3} model. The semantic gap tends to increase as the reasoning effort increases, even though accuracy is not notably affected. This again aligns with previous experiments, in which the semantic gap seems to grow as task accuracy increases, meaning that the better the model, the more fragile it is to semantic changes, as we measure in WhatCounts.

\begin{conclusionbox}{Reasoning Dependence}
    Reasoning does not change the semantic gap while not substantially improving counting accuracy.
\end{conclusionbox}

\section{How does Training Impact the Semantic Gap?}

While it is obvious that this semantic sensitivity is a result of the training, it is difficult to know which aspects of the training approach and data affect this problem, and to what extent. To study this, we evaluated a series of models trained on the same amount of data and with the same training specifications, starting from the same base model but using data from different datasets. For this, we evaluated the models from \citet{ivison2024unpacking}. In this work, they trained several models from the base model \texttt{Tulu2-13B} using the Reinforcement Learning (RL) techniques of DPO and PPO in 60K rows of the datasets of StackExchange \cite{h4stackexchange}, Nectar \cite{zhu2024starlingb}, and HH-RLHF \cite{bai2022training}. As these models have a smaller context window (4098) compared to the ones evaluated in the previous ablations, we reduced the count ranges in this experiment while keeping everything else the same (i.e., using the pipe as a separator, 20 questions per range, and the same six semantic classes considered).

\begin{figure}[h]
  \centering
  \includegraphics[width=\columnwidth]{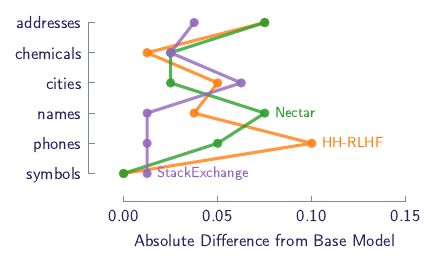}
  \caption{\textbf{Fine-tuning introduces highly variable performance across semantic classes.} Results are from experiments based on models fine-tuned on different datasets (shown in color), as a difference with respect to the results of the base model, Tulu-2-13B. The results show that training on only 60k instances of popular datasets leads to very different variations that are difficult to predict, as they do not follow a common trend across classes.}
  \label{fig:fig-finetunings}
\end{figure}

\begin{figure*}[!ht]
    \centering
    \includegraphics[width=\textwidth]{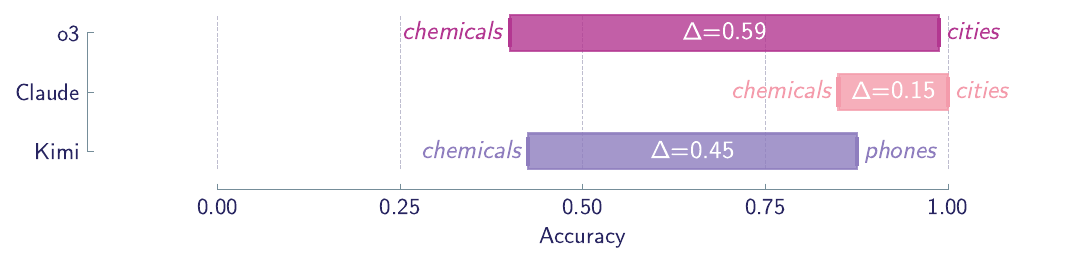}
    \caption{\textbf{Both accuracy and semantic gap increase for LLMs in an agentic setup.} Results are from experiments where models' agentic workflows are tasked with counting in a simulation of a batch system. Note that DeepSeek was not run for this experiment, as the provider that we used to run the model does not support tool calling.}
    \label{fig:agent-ablation}
\end{figure*}

\Cref{fig:fig-finetunings} shows the results of this evaluation as the absolute difference with respect to the base model \texttt{Tulu2-13B}. In the figure, only the models trained using DPO are shown, as we observed (see \Cref{tab:dpo-ppo-comparison}), the results of training with DPO or PPO did not differ substantially across the same datasets. In the figure, one can see that the training, despite being relatively small, has significant effects on the counting performance, even though the models have been trained on very different tasks. Additionally, it is possible to observe how the different datasets affect some entity classes, such as phone numbers, very differently. 

If we compare these findings with the variation observed for the same models in conventional benchmarks, such as in GSM8K shown in \Cref{tab:gsm8k-runs}, we observe that the differences in WhatCounts are substantially bigger. This sustains our observation that this semantic fragility is very difficult to control during training.

Also, the fact that the different training paradigms with the same data do not lead to differences means that what really matters for this fragility is the data itself.

\section{LLM Agents Inherit the Semantic Fragility}

To assess the impact of this task in the real world, we designed a specific agentic environment that simulates a length-guarded API. This is based on systems and servers that include integrity checks to detect partial or truncated requests, catch client-side bugs, or keep downstream systems consistent. Thus, the environment task prompt consists of a list of items. The agent is instructed to use the Python execution tool to count items, then submit the answer with the original list plus the count. Agents are constrained to five iterations, and the prompts and tools used are described in \Cref{sec:agent-prompts}, and in \Cref{sec:tools}

\Cref{fig:agent-ablation} shows the results when the agents are deployed in a batch system environment. While accuracy increases relative to the base models, so does the semantic gap. This suggests that the semantic gap also extends to real-world settings, where this semantic fragility influences how models use tools.
It is surprising that this effect is so strong that a trivial problem, such as this, that can be easily solved with the Python execution tool, is affected in such a strong way.


\section{Future Work}

A direct extension of this work is to increase the sample size and encompass more semantic classes. This would yield lower-variance estimators of the semantic gap, distinguishing more precisely between within-type and between-type variation. Such an expansion is supported by the modular design of our open-source codebase, which facilitates the integration of new semantic class generators and datasets.

On the theoretical side, existing formal algorithmic accounts, such as the RASP paradigm, are largely content-agnostic and thus cannot model argument-dependent semantic variation. A promising open direction is to extend such formalisms with mechanisms for semantic-conditioned computation. This line of research would help bridge the gap between abstract, symbolic descriptions of operators and our empirical observation that \enquote{LLM functions} are semantically sensitive.
\section{Conclusion}

In this work, we introduced WhatCounts, a controlled benchmark for evaluating the robustness of Large Language Models on a fundamental reasoning primitive: counting items in a list. Our central finding is unexpected: counting accuracy varies by over 40\% depending solely on what is being counted.

This is not the fragility we might anticipate. Prior work has shown that LLMs are sensitive to prompt phrasing, task complexity, list length, and tokenization artifacts. These are, in some sense, understandable failure modes---variations in how a task is presented or how hard it is. What we uncover is different: under identical formatting, identical list lengths, and identical instructions, models succeed or fail based on the semantic class of the arguments. Cities behave differently from chemicals. Names behave differently from symbols. The task is the same; only the meaning of the inputs changes.
Through systematic ablations, we ruled out surface explanations. Models correctly segment classes that they cannot count. Explicit structure does not help. Token-count controls reproduce the same patterns. 

These results force a consequential conclusion: LLMs do not implement algorithms. They approximate them, and the approximation is argument-dependent in ways that violate what it means to be an algorithm.
This matters beyond counting. Counting is among the simplest algorithmic primitives---if any operation should be invariant to argument semantics, it is this one. If even counting carries hidden dependencies on the meaning of the inputs, then any LLM \enquote{function} such as sorting, searching, extracting, or aggregating may do the same. For emerging agentic pipelines, where outputs from one component feed into another, such biases create risks of cascading and compounded errors \cite{chen2025prompt, kumar2025robustness}. Our agentic example shows that this might be the case.
The assumption underlying \enquote{prompt programming} is that a prompted algorithm will behave like an algorithm, i.e., indifferent to the semantic class of its arguments. WhatCounts shows this assumption is false. What you count changes whether you can count it.

\section{Acknowledgments}

This work was supported by the Carl-Zeiss Foundation.
K.M.J. is part of the NFDI consortium FAIRmat funded by the Deutsche Forschungsgemeinschaft (DFG, German Research Foundation) – project 460197019, and was also supported by a Google Research Scholar Award. 
Parts of the work were also supported by a gift by OpenPhilanthropy. 
We thank Meiling Sun, Mara Schilling-Wilhelmi, Gordan Prastalo, and Ali Asghar Aghajani for their feedback on a draft of this article.

\clearpage

\bibliography{references.bib}

@article{nezhurina2024alice,
  title   = {Alice in Wonderland: Simple Tasks Showing Complete Reasoning Breakdown in State-Of-the-Art Large Language Models},
  author  = {Marianna Nezhurina and Lucia Cipolina-Kun and Mehdi Cherti and Jenia Jitsev},
  year    = {2024},
  journal = {arXiv preprint arXiv: 2406.02061}
}

@article{mirzadeh2024gsm0symbolic0,
  title     = {GSM-Symbolic: Understanding the Limitations of Mathematical Reasoning in Large Language Models},
  author    = {Iman Mirzadeh and Keivan Alizadeh-Vahid and Hooman Shahrokhi and Oncel Tuzel and Samy Bengio and Mehrdad Farajtabar},
  journal   = {International Conference on Learning Representations},
  year      = {2024},
  bibSource = {Semantic Scholar https://www.semanticscholar.org/paper/05506581cade1a8ef6372616cec20b81a3d5c366}
}

@article{chen2025prompt,
  title   = {Prompt Stability Matters: Evaluating and Optimizing Auto-Generated Prompt in General-Purpose Systems},
  author  = {Ke Chen and Yufei Zhou and Xitong Zhang and Haohan Wang},
  year    = {2025},
  journal = {arXiv preprint arXiv: 2505.13546}
}

@article{kumar2025robustness,
  title     = {Robustness in Large Language Models: A Survey of Mitigation Strategies and Evaluation Metrics},
  author    = {Pankaj Kumar and Subhankar Mishra},
  journal   = {Trans. Mach. Learn. Res.},
  year      = {2025},
  doi       = {10.48550/arXiv.2505.18658},
}

@inproceedings{
sclar2024quantifying,
title={Quantifying Language Models' Sensitivity to Spurious Features in Prompt Design or: How I learned to start worrying about prompt formatting},
author={Melanie Sclar and Yejin Choi and Yulia Tsvetkov and Alane Suhr},
booktitle={The Twelfth International Conference on Learning Representations},
year={2024},
url={https://openreview.net/forum?id=RIu5lyNXjT}
}

@article{Alahmari2025large,
  title = {Large language models robustness against perturbation},
  volume = {16},
  ISSN = {2045-2322},
  url = {http://dx.doi.org/10.1038/s41598-025-29770-0},
  DOI = {10.1038/s41598-025-29770-0},
  number = {1},
  journal = {Scientific Reports},
  publisher = {Springer Science and Business Media LLC},
  author = {Alahmari,  Saeed S. and Hall,  Lawrence and Mouton,  Peter R. and Goldgof,  Dmitry},
  year = {2025},
  month = nov 
}

@article{haller2025llm,
  title   = {LLM Knowledge is Brittle: Truthfulness Representations Rely on Superficial Resemblance},
  author  = {Patrick Haller and Mark Ibrahim and Polina Kirichenko and Levent Sagun and Samuel J. Bell},
  year    = {2025},
  journal = {arXiv preprint arXiv: 2510.11905}
}

@article{lin2024llm,
  title   = {LLM Whisperer: An Inconspicuous Attack to Bias LLM Responses},
  author  = {Weiran Lin and Anna Gerchanovsky and Omer Akgul and Lujo Bauer and Matt Fredrikson and Zifan Wang},
  year    = {2024},
  journal = {arXiv preprint arXiv: 2406.04755}
}

@article{ullman2023large,
  title   = {Large Language Models Fail on Trivial Alterations to Theory-of-Mind Tasks},
  author  = {Tomer Ullman},
  year    = {2023},
  journal = {arXiv preprint arXiv: 2302.08399}
}

@inproceedings{errica2025what,
  author    = {Federico Errica and Davide Sanvito and Giuseppe Siracusano and Roberto Bifulco},
  editor    = {Luis Chiruzzo and Alan Ritter and Lu Wang},
  title     = {What Did {I} Do Wrong? Quantifying LLMs' Sensitivity and Consistency to Prompt Engineering},
  booktitle = {Proceedings of the 2025 Conference of the Nations of the Americas Chapter of the Association for Computational Linguistics: Human Language Technologies, {NAACL} 2025 - Volume 1: Long Papers, Albuquerque, New Mexico, USA, April 29 - May 4, 2025},
  pages     = {1543-1558},
  publisher = {Association for Computational Linguistics},
  year      = {2025},
  url       = {https://doi.org/10.18653/v1/2025.naacl-long.73},
  doi       = {10.18653/V1/2025.NAACL-LONG.73},
  bibsource = {dblp computer science bibliography, https://dblp.org}
}

@article{cobbe2021training,
  title   = {Training Verifiers to Solve Math Word Problems},
  author  = {Karl Cobbe and Vineet Kosaraju and Mohammad Bavarian and Mark Chen and Heewoo Jun and Lukasz Kaiser and Matthias Plappert and Jerry Tworek and Jacob Hilton and Reiichiro Nakano and Christopher Hesse and John Schulman},
  year    = {2021},
  journal = {arXiv preprint arXiv: 2110.14168}
}

@article{xu2025llm,
  title     = {LLM The Genius Paradox: A Linguistic and Math Expert's Struggle with Simple Word-based Counting Problems},
  author    = {Nan Xu and Xuezhe Ma},
  journal   = {North American Chapter of the Association for Computational Linguistics},
  year      = {2025},
  doi       = {10.18653/v1/2025.naacl-long.172},
  bibSource = {Semantic Scholar https://www.semanticscholar.org/paper/92743fca569cdad9ff0aa104eee661aba368a215}
}

@article{zhang2024counting,
  title   = {Counting Ability of Large Language Models and Impact of Tokenization},
  author  = {Xiang Zhang and Juntai Cao and Chenyu You},
  year    = {2024},
  journal = {arXiv preprint arXiv: 2410.19730}
}

@inproceedings{chang2025language,
  author    = {Yingshan Chang and Yonatan Bisk},
  title     = {Language Models Need Inductive Biases to Count Inductively},
  booktitle = {The Thirteenth International Conference on Learning Representations, {ICLR} 2025, Singapore, April 24-28, 2025},
  publisher = {OpenReview.net},
  year      = {2025},
  url       = {https://openreview.net/forum?id=s3IBHTTDYl},
  bibsource = {dblp computer science bibliography, https://dblp.org}
}

@article{fu2024large,
  title   = {Why Do Large Language Models (LLMs) Struggle to Count Letters?},
  author  = {Tairan Fu and Raquel Ferrando and Javier Conde and Carlos Arriaga and Pedro Reviriego},
  year    = {2024},
  journal = {arXiv preprint arXiv: 2412.18626}
}

@article{srivastava2022bigbench,
  author    = {Aarohi Srivastava and Abhinav Rastogi and Abhishek Rao and Abu Awal Md Shoeb and Abubakar Abid and Adam Fisch and Adam R. Brown and Adam Santoro and Aditya Gupta and Adri{\`{a}} Garriga{-}Alonso and Agnieszka Kluska and Aitor Lewkowycz and Akshat Agarwal and Alethea Power and Alex Ray and Alex Warstadt and Alexander W. Kocurek and Ali Safaya and Ali Tazarv and Alice Xiang and Alicia Parrish and Allen Nie and Aman Hussain and Amanda Askell and Amanda Dsouza and Ambrose Slone and Ameet Rahane and Anantharaman S. Iyer and Anders Andreassen and Andrea Madotto and Andrea Santilli and Andreas Stuhlm{\"{u}}ller and Andrew M. Dai and Andrew La and Andrew K. Lampinen and Andy Zou and Angela Jiang and Angelica Chen and Anh Vuong and Animesh Gupta and Anna Gottardi and Antonio Norelli and Anu Venkatesh and Arash Gholamidavoodi and Arfa Tabassum and Arul Menezes and Arun Kirubarajan and Asher Mullokandov and Ashish Sabharwal and Austin Herrick and Avia Efrat and Aykut Erdem and Ayla Karakas and B. Ryan Roberts and Bao Sheng Loe and Barret Zoph and Bartlomiej Bojanowski and Batuhan {\"{O}}zyurt and Behnam Hedayatnia and Behnam Neyshabur and Benjamin Inden and Benno Stein and Berk Ekmekci and Bill Yuchen Lin and Blake Howald and Bryan Orinion and Cameron Diao and Cameron Dour and Catherine Stinson and Cedrick Argueta and C{\`{e}}sar Ferri Ram{\'{\i}}rez and Chandan Singh and Charles Rathkopf and Chenlin Meng and Chitta Baral and Chiyu Wu and Chris Callison{-}Burch and Chris Waites and Christian Voigt and Christopher D. Manning and Christopher Potts and Cindy Ramirez and Clara E. Rivera and Clemencia Siro and Colin Raffel and Courtney Ashcraft and Cristina Garbacea and Damien Sileo and Dan Garrette and Dan Hendrycks and Dan Kilman and Dan Roth and Daniel Freeman and Daniel Khashabi and Daniel Levy and Daniel Mosegu{\'{\i}} Gonz{\'{a}}lez and Danielle Perszyk and Danny Hernandez and Danqi Chen and Daphne Ippolito and Dar Gilboa and David Dohan and David Drakard and David Jurgens and Debajyoti Datta and Deep Ganguli and Denis Emelin and Denis Kleyko and Deniz Yuret and Derek Chen and Derek Tam and Dieuwke Hupkes and Diganta Misra and Dilyar Buzan and Dimitri Coelho Mollo and Diyi Yang and Dong{-}Ho Lee and Dylan Schrader and Ekaterina Shutova and Ekin Dogus Cubuk and Elad Segal and Eleanor Hagerman and Elizabeth Barnes and Elizabeth Donoway and Ellie Pavlick and Emanuele Rodol{\`{a}} and Emma Lam and Eric Chu and Eric Tang and Erkut Erdem and Ernie Chang and Ethan A. Chi and Ethan Dyer and Ethan J. Jerzak and Ethan Kim and Eunice Engefu Manyasi and Evgenii Zheltonozhskii and Fanyue Xia and Fatemeh Siar and Fernando Mart{\'{\i}}nez{-}Plumed and Francesca Happ{\'{e}} and Fran{\c{c}}ois Chollet and Frieda Rong and Gaurav Mishra and Genta Indra Winata and Gerard de Melo and Germ{\'{a}}n Kruszewski and Giambattista Parascandolo and Giorgio Mariani and Gloria Wang and Gonzalo Jaimovitch{-}L{\'{o}}pez and Gregor Betz and Guy Gur{-}Ari and Hana Galijasevic and Hannah Kim and Hannah Rashkin and Hannaneh Hajishirzi and Harsh Mehta and Hayden Bogar and Henry Shevlin and Hinrich Sch{\"{u}}tze and Hiromu Yakura and Hongming Zhang and Hugh Mee Wong and Ian Ng and Isaac Noble and Jaap Jumelet and Jack Geissinger and Jackson Kernion and Jacob Hilton and Jaehoon Lee and Jaime Fern{\'{a}}ndez Fisac and James B. Simon and James Koppel and James Zheng and James Zou and Jan Kocon and Jana Thompson and Janelle Wingfield and Jared Kaplan and Jarema Radom and Jascha Sohl{-}Dickstein and Jason Phang and Jason Wei and Jason Yosinski and Jekaterina Novikova and Jelle Bosscher and Jennifer Marsh and Jeremy Kim and Jeroen Taal and Jesse H. Engel and Jesujoba Alabi and Jiacheng Xu and Jiaming Song and Jillian Tang and Joan Waweru and John Burden and John Miller and John U. Balis and Jonathan Batchelder and Jonathan Berant and J{\"{o}}rg Frohberg and Jos Rozen and Jos{\'{e}} Hern{\'{a}}ndez{-}Orallo and Joseph Boudeman and Joseph Guerr and Joseph Jones and Joshua B. Tenenbaum and Joshua S. Rule and Joyce Chua and Kamil Kanclerz and Karen Livescu and Karl Krauth and Karthik Gopalakrishnan and Katerina Ignatyeva and Katja Markert and Kaustubh D. Dhole and Kevin Gimpel and Kevin Omondi and Kory W. Mathewson and Kristen Chiafullo and Ksenia Shkaruta and Kumar Shridhar and Kyle McDonell and Kyle Richardson and Laria Reynolds and Leo Gao and Li Zhang and Liam Dugan and Lianhui Qin and Lidia Contreras Ochando and Louis{-}Philippe Morency and Luca Moschella and Lucas Lam and Lucy Noble and Ludwig Schmidt and Luheng He and Luis Oliveros Col{\'{o}}n and Luke Metz and L{\"{u}}tfi Kerem Senel and Maarten Bosma and Maarten Sap and Maartje ter Hoeve and Maheen Farooqi and Manaal Faruqui and Mantas Mazeika and Marco Baturan and Marco Marelli and Marco Maru and Mar{\'{\i}}a Jos{\'{e}} Ram{\'{\i}}rez{-}Quintana and Marie Tolkiehn and Mario Giulianelli and Martha Lewis and Martin Potthast and Matthew L. Leavitt and Matthias Hagen and M{\'{a}}ty{\'{a}}s Schubert and Medina Baitemirova and Melody Arnaud and Melvin McElrath and Michael A. Yee and Michael Cohen and Michael Gu and Michael I. Ivanitskiy and Michael Starritt and Michael Strube and Michal Swedrowski and Michele Bevilacqua and Michihiro Yasunaga and Mihir Kale and Mike Cain and Mimee Xu and Mirac Suzgun and Mitch Walker and Mo Tiwari and Mohit Bansal and Moin Aminnaseri and Mor Geva and Mozhdeh Gheini and Mukund Varma T. and Nanyun Peng and Nathan A. Chi and Nayeon Lee and Neta Gur{-}Ari Krakover and Nicholas Cameron and Nicholas Roberts and Nick Doiron and Nicole Martinez and Nikita Nangia and Niklas Deckers and Niklas Muennighoff and Nitish Shirish Keskar and Niveditha Iyer and Noah Constant and Noah Fiedel and Nuan Wen and Oliver Zhang and Omar Agha and Omar Elbaghdadi and Omer Levy and Owain Evans and Pablo Antonio Moreno Casares and Parth Doshi and Pascale Fung and Paul Pu Liang and Paul Vicol and Pegah Alipoormolabashi and Peiyuan Liao and Percy Liang and Peter Chang and Peter Eckersley and Phu Mon Htut and Pinyu Hwang and Piotr Milkowski and Piyush Patil and Pouya Pezeshkpour and Priti Oli and Qiaozhu Mei and Qing Lyu and Qinlang Chen and Rabin Banjade and Rachel Etta Rudolph and Raefer Gabriel and Rahel Habacker and Ramon Risco and Rapha{\"{e}}l Milli{\`{e}}re and Rhythm Garg and Richard Barnes and Rif A. Saurous and Riku Arakawa and Robbe Raymaekers and Robert Frank and Rohan Sikand and Roman Novak and Roman Sitelew and Ronan LeBras and Rosanne Liu and Rowan Jacobs and Rui Zhang and Ruslan Salakhutdinov and Ryan Chi and Ryan Lee and Ryan Stovall and Ryan Teehan and Rylan Yang and Sahib Singh and Saif M. Mohammad and Sajant Anand and Sam Dillavou and Sam Shleifer and Sam Wiseman and Samuel Gruetter and Samuel R. Bowman and Samuel S. Schoenholz and Sanghyun Han and Sanjeev Kwatra and Sarah A. Rous and Sarik Ghazarian and Sayan Ghosh and Sean Casey and Sebastian Bischoff and Sebastian Gehrmann and Sebastian Schuster and Sepideh Sadeghi and Shadi Hamdan and Sharon Zhou and Shashank Srivastava and Sherry Shi and Shikhar Singh and Shima Asaadi and Shixiang Shane Gu and Shubh Pachchigar and Shubham Toshniwal and Shyam Upadhyay and Shyamolima (Shammie) Debnath and Siamak Shakeri and Simon Thormeyer and Simone Melzi and Siva Reddy and Sneha Priscilla Makini and Soo{-}Hwan Lee and Spencer Torene and Sriharsha Hatwar and Stanislas Dehaene and Stefan Divic and Stefano Ermon and Stella Biderman and Stephanie Lin and Stephen Prasad and Steven T. Piantadosi and Stuart M. Shieber and Summer Misherghi and Svetlana Kiritchenko and Swaroop Mishra and Tal Linzen and Tal Schuster and Tao Li and Tao Yu and Tariq Ali and Tatsu Hashimoto and Te{-}Lin Wu and Th{\'{e}}o Desbordes and Theodore Rothschild and Thomas Phan and Tianle Wang and Tiberius Nkinyili and Timo Schick and Timofei Kornev and Titus Tunduny and Tobias Gerstenberg and Trenton Chang and Trishala Neeraj and Tushar Khot and Tyler Shultz and Uri Shaham and Vedant Misra and Vera Demberg and Victoria Nyamai and Vikas Raunak and Vinay V. Ramasesh and Vinay Uday Prabhu and Vishakh Padmakumar and Vivek Srikumar and William Fedus and William Saunders and William Zhang and Wout Vossen and Xiang Ren and Xiaoyu Tong and Xinran Zhao and Xinyi Wu and Xudong Shen and Yadollah Yaghoobzadeh and Yair Lakretz and Yangqiu Song and Yasaman Bahri and Yejin Choi and Yichi Yang and Yiding Hao and Yifu Chen and Yonatan Belinkov and Yu Hou and Yufang Hou and Yuntao Bai and Zachary Seid and Zhuoye Zhao and Zijian Wang and Zijie J. Wang and Zirui Wang and Ziyi Wu},
  title     = {Beyond the Imitation Game: Quantifying and extrapolating the capabilities of language models},
  journal   = {Trans. Mach. Learn. Res.},
  volume    = {2023},
  year      = {2023},
  url       = {https://openreview.net/forum?id=uyTL5Bvosj},
  bibsource = {dblp computer science bibliography, https://dblp.org}
}

@article{ball2024canwe,
  author    = {Thomas Ball and Shuo Chen and Cormac Herley},
  title     = {Can We Count on LLMs? The Fixed-Effect Fallacy and Claims of {GPT-4} Capabilities},
  journal   = {Trans. Mach. Learn. Res.},
  volume    = {2024},
  year      = {2024},
  url       = {https://openreview.net/forum?id=qt4d0EGZsK},
  bibsource = {dblp computer science bibliography, https://dblp.org}
}

@article{yehudai2024transformers,
  title   = {When Can Transformers Count to n?},
  author  = {Gilad Yehudai and Haim Kaplan and Asma Ghandeharioun and Mor Geva and Amir Globerson},
  year    = {2024},
  journal = {arXiv preprint arXiv: 2407.15160}
}

@article{liu2025evaluating,
  title   = {Evaluating Robustness of Large Language Models Against Multilingual Typographical Errors},
  author  = {Yihong Liu and Raoyuan Zhao and Lena Altinger and Hinrich Schütze and Michael A. Hedderich},
  year    = {2025},
  journal = {arXiv preprint arXiv: 2510.09536}
}

@article{cosma2025strawberry,
  title   = {The Strawberry Problem: Emergence of Character-level Understanding in Tokenized Language Models},
  author  = {Adrian Cosma and Stefan Ruseti and Emilian Radoi and Mihai Dascalu},
  year    = {2025},
  journal = {arXiv preprint arXiv: 2505.14172}
}

@article{goldblum2023free,
  title   = {The No Free Lunch Theorem, Kolmogorov Complexity, and the Role of Inductive Biases in Machine Learning},
  author  = {Micah Goldblum and Marc Finzi and Keefer Rowan and Andrew Gordon Wilson},
  year    = {2023},
  journal = {arXiv preprint arXiv: 2304.05366}
}

@article{ivison2024unpacking,
  title   = {Unpacking DPO and PPO: Disentangling Best Practices for Learning from Preference Feedback},
  author  = {Hamish Ivison and Yizhong Wang and Jiacheng Liu and Zeqiu Wu and Valentina Pyatkin and Nathan Lambert and Noah A. Smith and Yejin Choi and Hannaneh Hajishirzi},
  year    = {2024},
  journal = {arXiv preprint arXiv: 2406.09279}
}

@inproceedings{
zhu2024starlingb,
title={Starling-7B: Improving Helpfulness and Harmlessness with {RLAIF}},
author={Banghua Zhu and Evan Frick and Tianhao Wu and Hanlin Zhu and Karthik Ganesan and Wei-Lin Chiang and Jian Zhang and Jiantao Jiao},
booktitle={First Conference on Language Modeling},
year={2024},
url={https://openreview.net/forum?id=GqDntYTTbk}
}

@article{bai2022training,
  title   = {Training a Helpful and Harmless Assistant with Reinforcement Learning from Human Feedback},
  author  = {Yuntao Bai and Andy Jones and Kamal Ndousse and Amanda Askell and Anna Chen and Nova DasSarma and Dawn Drain and Stanislav Fort and Deep Ganguli and Tom Henighan and Nicholas Joseph and Saurav Kadavath and Jackson Kernion and Tom Conerly and Sheer El-Showk and Nelson Elhage and Zac Hatfield-Dodds and Danny Hernandez and Tristan Hume and Scott Johnston and Shauna Kravec and Liane Lovitt and Neel Nanda and Catherine Olsson and Dario Amodei and Tom Brown and Jack Clark and Sam McCandlish and Chris Olah and Ben Mann and Jared Kaplan},
  year    = {2022},
  journal = {arXiv preprint arXiv: 2204.05862}
}

@misc{h4stackexchange,
  author       = {Lambert, Nathan and Tunstall, Lewis and Rajani, Nazneen and Thrush, Tristan},
  title        = {HuggingFace H4 Stack Exchange Preference Dataset},
  year         = {2023},
  howpublished = {\url{https://huggingface.co/datasets/HuggingFaceH4/stack-exchange-preferences}},
  note         = {Hugging Face dataset. Accessed: 2026-01-21}
}

@inproceedings{weiss2021thinking,
  author    = {Gail Weiss and Yoav Goldberg and Eran Yahav},
  editor    = {Marina Meila and Tong Zhang},
  title     = {Thinking Like Transformers},
  booktitle = {Proceedings of the 38th International Conference on Machine Learning, {ICML} 2021, 18-24 July 2021, Virtual Event},
  series    = {Proceedings of Machine Learning Research},
  volume    = {139},
  pages     = {11080-11090},
  publisher = {PMLR},
  year      = {2021},
  url       = {http://proceedings.mlr.press/v139/weiss21a.html},
  bibsource = {dblp computer science bibliography, https://dblp.org}
}

@article{merrill2024expressive,
  title     = {The Expressive Power of Transformers with Chain of Thought},
  author    = {William Merrill and Ashish Sabharwal},
  journal   = {International Conference on Learning Representations},
  year      = {2024},
  doi       = {10.48550/arXiv.2310.07923},
  bibSource = {Semantic Scholar https://www.semanticscholar.org/paper/75c19f3249f644f5cb2182282fc117c089fd3f65}
}

@article{chiang2024transformers,
  title   = {Transformers in Uniform TC$^0$},
  author  = {David Chiang},
  year    = {2024},
  journal = {arXiv preprint arXiv: 2409.13629}
}

@inproceedings{chiang2023tighter,
  title     = {Tighter Bounds on the Expressivity of Transformer Encoders},
  author    = {Chiang, David and Cholak, Peter and Pillay, Anand},
  booktitle = {Proceedings of the 40th International Conference on Machine Learning},
  pages     = {5544-5562},
  year      = {2023},
  editor    = {Krause, Andreas and Brunskill, Emma and Cho, Kyunghyun and Engelhardt, Barbara and Sabato, Sivan and Scarlett, Jonathan},
  volume    = {202},
  series    = {Proceedings of Machine Learning Research},
  month     = {23-29 Jul},
  publisher = {PMLR},
  url       = {https://proceedings.mlr.press/v202/chiang23a.html}
}

@software{faker_pypi_2026,
  title        = {Faker: Python package that generates fake data},
  author       = {joke2k},
  version      = {40.1.2},
  date         = {2026-01-13},
  url          = {https://pypi.org/project/Faker/},
  urldate      = {2026-01-26},
  year         = {2026},
  note         = {Source code: https://github.com/joke2k/faker}
}

@article{mirza2025chempile0,
  title   = {ChemPile: A 250GB Diverse and Curated Dataset for Chemical Foundation Models},
  author  = {Adrian Mirza and Nawaf Alampara and Martiño Ríos-García and Mohamed Abdelalim and Jack Butler and Bethany Connolly and Tunca Dogan and Marianna Nezhurina and Bünyamin Şen and Santosh Tirunagari and Mark Worrall and Adamo Young and Philippe Schwaller and Michael Pieler and Kevin Maik Jablonka},
  year    = {2025},
  journal = {arXiv preprint arXiv: 2505.12534}
}

@misc{coolsymbol_all_symbols_2025,
  title        = {All Cool Symbol Copy and Paste},
  howpublished = {CoolSymbol.top (web page)},
  year         = {2025},
  url          = {https://www.coolsymbol.top/all-symbol.html},
  urldate      = {2026-01-26},
  note         = {Copyright notice on page: © 2025.}
}

@misc{geonames_all_cities_pop1000_opendatasoft_2025,
  title        = {Geonames - All Cities with a population > 1000},
  author       = {{GeoNames}},
  year         = {2025},
  howpublished = {Opendatasoft Public Data portal (dataset table view)},
  url          = {https://public.opendatasoft.com/explore/dataset/geonames-all-cities-with-a-population-1000/table/?disjunctive.cou_name_en&sort=cou_name_en},
  urldate      = {2026-01-26},
  note         = {Dataset identifier: geonames-all-cities-with-a-population-1000. Publisher: GeoNames. License: CC BY 4.0. Last modified: 2025-12-09 (portal metadata).}
}

@article{ouellette2023counting,
  title   = {Counting and Algorithmic Generalization with Transformers},
  author  = {Simon Ouellette and Rolf Pfister and Hansueli Jud},
  year    = {2023},
  journal = {arXiv preprint arXiv: 2310.08661}
}

@inproceedings{zhou2024what,
  author    = {Hattie Zhou and Arwen Bradley and Etai Littwin and Noam Razin and Omid Saremi and Joshua M. Susskind and Samy Bengio and Preetum Nakkiran},
  title     = {What Algorithms can Transformers Learn? {A} Study in Length Generalization},
  booktitle = {The Twelfth International Conference on Learning Representations, {ICLR} 2024, Vienna, Austria, May 7-11, 2024},
  publisher = {OpenReview.net},
  year      = {2024},
  url       = {https://openreview.net/forum?id=AssIuHnmHX},
  bibsource = {dblp computer science bibliography, https://dblp.org}
}

@article{liang2025prompts,
  title     = {Prompts are programs too! understanding how developers build software containing prompts},
  author    = {Liang, Jenny T and Lin, Melissa and Rao, Nikitha and Myers, Brad A},
  journal   = {Proceedings of the ACM on Software Engineering},
  volume    = {2},
  number    = {FSE},
  pages     = {1591-1614},
  year      = {2025},
  publisher = {ACM New York, NY, USA}
}

@article{reynolds2021prompt,
  title   = {Prompt Programming for Large Language Models: Beyond the Few-Shot Paradigm},
  author  = {Laria Reynolds and Kyle McDonell},
  year    = {2021},
  journal = {arXiv preprint arXiv: 2102.07350}
}

@article{schick2023toolformer,
  title   = {Toolformer: Language models can teach themselves to use tools},
  author  = {Schick, Timo and Dwivedi-Yu, Jane and Dess{\`\i}, Roberto and Raileanu, Roberta and Lomeli, Maria and Hambro, Eric and Zettlemoyer, Luke and Cancedda, Nicola and Scialom, Thomas},
  journal = {Advances in Neural Information Processing Systems},
  volume  = {36},
  pages   = {68539-68551},
  year    = {2023}
}

@article{herbold2025sortbench0,
  title   = {SortBench: Benchmarking LLMs based on their ability to sort lists},
  author  = {Steffen Herbold},
  year    = {2025},
  journal = {arXiv preprint arXiv: 2504.08312}
}

@article{suzgun2022challenging,
  title     = {Challenging BIG-Bench Tasks and Whether Chain-of-Thought Can Solve Them},
  author    = {Mirac Suzgun and Nathan Scales and Nathanael Scharli and Sebastian Gehrmann and Yi Tay and Hyung Won Chung and A. Chowdhery and Quoc V. Le and Ed H. Chi and Denny Zhou and Jason Wei},
  journal   = {Annual Meeting of the Association for Computational Linguistics},
  year      = {2022},
  doi       = {10.48550/arXiv.2210.09261},
  bibSource = {Semantic Scholar https://www.semanticscholar.org/paper/663a41c866d49ce052801fbc88947d39764cad29}
}

@article{srivastave2023beyond,
  author    = {Aarohi Srivastava and Abhinav Rastogi and Abhishek Rao and Abu Awal Md Shoeb and Abubakar Abid and Adam Fisch and Adam R. Brown and Adam Santoro and Aditya Gupta and Adri{\`{a}} Garriga{-}Alonso and Agnieszka Kluska and Aitor Lewkowycz and Akshat Agarwal and Alethea Power and Alex Ray and Alex Warstadt and Alexander W. Kocurek and Ali Safaya and Ali Tazarv and Alice Xiang and Alicia Parrish and Allen Nie and Aman Hussain and Amanda Askell and Amanda Dsouza and Ambrose Slone and Ameet Rahane and Anantharaman S. Iyer and Anders Andreassen and Andrea Madotto and Andrea Santilli and Andreas Stuhlm{\"{u}}ller and Andrew M. Dai and Andrew La and Andrew K. Lampinen and Andy Zou and Angela Jiang and Angelica Chen and Anh Vuong and Animesh Gupta and Anna Gottardi and Antonio Norelli and Anu Venkatesh and Arash Gholamidavoodi and Arfa Tabassum and Arul Menezes and Arun Kirubarajan and Asher Mullokandov and Ashish Sabharwal and Austin Herrick and Avia Efrat and Aykut Erdem and Ayla Karakas and B. Ryan Roberts and Bao Sheng Loe and Barret Zoph and Bartlomiej Bojanowski and Batuhan {\"{O}}zyurt and Behnam Hedayatnia and Behnam Neyshabur and Benjamin Inden and Benno Stein and Berk Ekmekci and Bill Yuchen Lin and Blake Howald and Bryan Orinion and Cameron Diao and Cameron Dour and Catherine Stinson and Cedrick Argueta and C{\`{e}}sar Ferri Ram{\'{\i}}rez and Chandan Singh and Charles Rathkopf and Chenlin Meng and Chitta Baral and Chiyu Wu and Chris Callison{-}Burch and Chris Waites and Christian Voigt and Christopher D. Manning and Christopher Potts and Cindy Ramirez and Clara E. Rivera and Clemencia Siro and Colin Raffel and Courtney Ashcraft and Cristina Garbacea and Damien Sileo and Dan Garrette and Dan Hendrycks and Dan Kilman and Dan Roth and Daniel Freeman and Daniel Khashabi and Daniel Levy and Daniel Mosegu{\'{\i}} Gonz{\'{a}}lez and Danielle Perszyk and Danny Hernandez and Danqi Chen and Daphne Ippolito and Dar Gilboa and David Dohan and David Drakard and David Jurgens and Debajyoti Datta and Deep Ganguli and Denis Emelin and Denis Kleyko and Deniz Yuret and Derek Chen and Derek Tam and Dieuwke Hupkes and Diganta Misra and Dilyar Buzan and Dimitri Coelho Mollo and Diyi Yang and Dong{-}Ho Lee and Dylan Schrader and Ekaterina Shutova and Ekin Dogus Cubuk and Elad Segal and Eleanor Hagerman and Elizabeth Barnes and Elizabeth Donoway and Ellie Pavlick and Emanuele Rodol{\`{a}} and Emma Lam and Eric Chu and Eric Tang and Erkut Erdem and Ernie Chang and Ethan A. Chi and Ethan Dyer and Ethan J. Jerzak and Ethan Kim and Eunice Engefu Manyasi and Evgenii Zheltonozhskii and Fanyue Xia and Fatemeh Siar and Fernando Mart{\'{\i}}nez{-}Plumed and Francesca Happ{\'{e}} and Fran{\c{c}}ois Chollet and Frieda Rong and Gaurav Mishra and Genta Indra Winata and Gerard de Melo and Germ{\'{a}}n Kruszewski and Giambattista Parascandolo and Giorgio Mariani and Gloria Wang and Gonzalo Jaimovitch{-}L{\'{o}}pez and Gregor Betz and Guy Gur{-}Ari and Hana Galijasevic and Hannah Kim and Hannah Rashkin and Hannaneh Hajishirzi and Harsh Mehta and Hayden Bogar and Henry Shevlin and Hinrich Sch{\"{u}}tze and Hiromu Yakura and Hongming Zhang and Hugh Mee Wong and Ian Ng and Isaac Noble and Jaap Jumelet and Jack Geissinger and Jackson Kernion and Jacob Hilton and Jaehoon Lee and Jaime Fern{\'{a}}ndez Fisac and James B. Simon and James Koppel and James Zheng and James Zou and Jan Kocon and Jana Thompson and Janelle Wingfield and Jared Kaplan and Jarema Radom and Jascha Sohl{-}Dickstein and Jason Phang and Jason Wei and Jason Yosinski and Jekaterina Novikova and Jelle Bosscher and Jennifer Marsh and Jeremy Kim and Jeroen Taal and Jesse H. Engel and Jesujoba Alabi and Jiacheng Xu and Jiaming Song and Jillian Tang and Joan Waweru and John Burden and John Miller and John U. Balis and Jonathan Batchelder and Jonathan Berant and J{\"{o}}rg Frohberg and Jos Rozen and Jos{\'{e}} Hern{\'{a}}ndez{-}Orallo and Joseph Boudeman and Joseph Guerr and Joseph Jones and Joshua B. Tenenbaum and Joshua S. Rule and Joyce Chua and Kamil Kanclerz and Karen Livescu and Karl Krauth and Karthik Gopalakrishnan and Katerina Ignatyeva and Katja Markert and Kaustubh D. Dhole and Kevin Gimpel and Kevin Omondi and Kory W. Mathewson and Kristen Chiafullo and Ksenia Shkaruta and Kumar Shridhar and Kyle McDonell and Kyle Richardson and Laria Reynolds and Leo Gao and Li Zhang and Liam Dugan and Lianhui Qin and Lidia Contreras Ochando and Louis{-}Philippe Morency and Luca Moschella and Lucas Lam and Lucy Noble and Ludwig Schmidt and Luheng He and Luis Oliveros Col{\'{o}}n and Luke Metz and L{\"{u}}tfi Kerem Senel and Maarten Bosma and Maarten Sap and Maartje ter Hoeve and Maheen Farooqi and Manaal Faruqui and Mantas Mazeika and Marco Baturan and Marco Marelli and Marco Maru and Mar{\'{\i}}a Jos{\'{e}} Ram{\'{\i}}rez{-}Quintana and Marie Tolkiehn and Mario Giulianelli and Martha Lewis and Martin Potthast and Matthew L. Leavitt and Matthias Hagen and M{\'{a}}ty{\'{a}}s Schubert and Medina Baitemirova and Melody Arnaud and Melvin McElrath and Michael A. Yee and Michael Cohen and Michael Gu and Michael I. Ivanitskiy and Michael Starritt and Michael Strube and Michal Swedrowski and Michele Bevilacqua and Michihiro Yasunaga and Mihir Kale and Mike Cain and Mimee Xu and Mirac Suzgun and Mitch Walker and Mo Tiwari and Mohit Bansal and Moin Aminnaseri and Mor Geva and Mozhdeh Gheini and Mukund Varma T. and Nanyun Peng and Nathan A. Chi and Nayeon Lee and Neta Gur{-}Ari Krakover and Nicholas Cameron and Nicholas Roberts and Nick Doiron and Nicole Martinez and Nikita Nangia and Niklas Deckers and Niklas Muennighoff and Nitish Shirish Keskar and Niveditha Iyer and Noah Constant and Noah Fiedel and Nuan Wen and Oliver Zhang and Omar Agha and Omar Elbaghdadi and Omer Levy and Owain Evans and Pablo Antonio Moreno Casares and Parth Doshi and Pascale Fung and Paul Pu Liang and Paul Vicol and Pegah Alipoormolabashi and Peiyuan Liao and Percy Liang and Peter Chang and Peter Eckersley and Phu Mon Htut and Pinyu Hwang and Piotr Milkowski and Piyush Patil and Pouya Pezeshkpour and Priti Oli and Qiaozhu Mei and Qing Lyu and Qinlang Chen and Rabin Banjade and Rachel Etta Rudolph and Raefer Gabriel and Rahel Habacker and Ramon Risco and Rapha{\"{e}}l Milli{\`{e}}re and Rhythm Garg and Richard Barnes and Rif A. Saurous and Riku Arakawa and Robbe Raymaekers and Robert Frank and Rohan Sikand and Roman Novak and Roman Sitelew and Ronan LeBras and Rosanne Liu and Rowan Jacobs and Rui Zhang and Ruslan Salakhutdinov and Ryan Chi and Ryan Lee and Ryan Stovall and Ryan Teehan and Rylan Yang and Sahib Singh and Saif M. Mohammad and Sajant Anand and Sam Dillavou and Sam Shleifer and Sam Wiseman and Samuel Gruetter and Samuel R. Bowman and Samuel S. Schoenholz and Sanghyun Han and Sanjeev Kwatra and Sarah A. Rous and Sarik Ghazarian and Sayan Ghosh and Sean Casey and Sebastian Bischoff and Sebastian Gehrmann and Sebastian Schuster and Sepideh Sadeghi and Shadi Hamdan and Sharon Zhou and Shashank Srivastava and Sherry Shi and Shikhar Singh and Shima Asaadi and Shixiang Shane Gu and Shubh Pachchigar and Shubham Toshniwal and Shyam Upadhyay and Shyamolima (Shammie) Debnath and Siamak Shakeri and Simon Thormeyer and Simone Melzi and Siva Reddy and Sneha Priscilla Makini and Soo{-}Hwan Lee and Spencer Torene and Sriharsha Hatwar and Stanislas Dehaene and Stefan Divic and Stefano Ermon and Stella Biderman and Stephanie Lin and Stephen Prasad and Steven T. Piantadosi and Stuart M. Shieber and Summer Misherghi and Svetlana Kiritchenko and Swaroop Mishra and Tal Linzen and Tal Schuster and Tao Li and Tao Yu and Tariq Ali and Tatsu Hashimoto and Te{-}Lin Wu and Th{\'{e}}o Desbordes and Theodore Rothschild and Thomas Phan and Tianle Wang and Tiberius Nkinyili and Timo Schick and Timofei Kornev and Titus Tunduny and Tobias Gerstenberg and Trenton Chang and Trishala Neeraj and Tushar Khot and Tyler Shultz and Uri Shaham and Vedant Misra and Vera Demberg and Victoria Nyamai and Vikas Raunak and Vinay V. Ramasesh and Vinay Uday Prabhu and Vishakh Padmakumar and Vivek Srikumar and William Fedus and William Saunders and William Zhang and Wout Vossen and Xiang Ren and Xiaoyu Tong and Xinran Zhao and Xinyi Wu and Xudong Shen and Yadollah Yaghoobzadeh and Yair Lakretz and Yangqiu Song and Yasaman Bahri and Yejin Choi and Yichi Yang and Yiding Hao and Yifu Chen and Yonatan Belinkov and Yu Hou and Yufang Hou and Yuntao Bai and Zachary Seid and Zhuoye Zhao and Zijian Wang and Zijie J. Wang and Zirui Wang and Ziyi Wu},
  title     = {Beyond the Imitation Game: Quantifying and extrapolating the capabilities of language models},
  journal   = {Trans. Mach. Learn. Res.},
  volume    = {2023},
  year      = {2023},
  url       = {https://openreview.net/forum?id=uyTL5Bvosj},
  bibsource = {dblp computer science bibliography, https://dblp.org}
}

@inproceedings{gao2023pal,
  author    = {Luyu Gao and Aman Madaan and Shuyan Zhou and Uri Alon and Pengfei Liu and Yiming Yang and Jamie Callan and Graham Neubig},
  editor    = {Andreas Krause and Emma Brunskill and Kyunghyun Cho and Barbara Engelhardt and Sivan Sabato and Jonathan Scarlett},
  title     = {{PAL:} Program-aided Language Models},
  booktitle = {International Conference on Machine Learning, {ICML} 2023, 23-29 July 2023, Honolulu, Hawaii, {USA}},
  series    = {Proceedings of Machine Learning Research},
  volume    = {202},
  pages     = {10764-10799},
  publisher = {PMLR},
  year      = {2023},
  url       = {https://proceedings.mlr.press/v202/gao23f.html},
  bibsource = {dblp computer science bibliography, https://dblp.org}
}
\bibliographystyle{icml2026}
\clearpage

\renewcommand\thefigure{\thesection.\arabic{figure}}

\renewcommand\thetable{\thesection.\arabic{table}}
\renewcommand\theequation{\thesection\arabic{equation}}
\setcounter{figure}{0}
\setcounter{table}{0}
\setcounter{equation}{0}

\appendix
\onecolumn
\section{Question Creation} \label{sec:obtain-entity}

For each semantic class, we select four different ranges: (7, 17), (85, 115), or (550, 650). For each range, 20 different questions are created for the same counting task. The list of inputs is created by sampling a random number of items from the corresponding range for each experiment without allowing duplicates. Then, the tasks are built by filling in the prompts for each task, as detailed in \Cref{sec:prompts}.

Each semantic class has its own factory in the code, and each factory samples randomly using different methods for that semantic class. Chemicals, cities, and symbols are randomly sampled from in-house databases. The chemicals database was created by taking all IUPAC names present in the ChemPile-mLIFT dataset \cite{mirza2025chempile0}. 
Cities and symbols were mined from dedicated webpages \cite{geonames_all_cities_pop1000_opendatasoft_2025, coolsymbol_all_symbols_2025}. 
Addresses, names, and phone numbers are generated using the Python package Faker \cite{faker_pypi_2026}. Faker manages its own random sampling, but it's initialized only once to avoid not actually doing random sampling when it's initialized for each group of questions. Faker allows us to choose from different locations for the three variables and between genders for the names. For the results reported in the paper, the items were sampled without regard to any of them, and mixing was performed across the following locations: \texttt{["en\_US", "en\_GB", "es\_ES", "fr\_FR", "de\_DE", "ja\_JP", "zh\_CN"]}. We also verified if some bias existed for one location or gender (results are detailed in \Cref{sec:bias-analysis}).  
The seed for all the random processes is set to 42, including the one used by Faker.

\section{Initial Observations}

\subsection{Answer as Round Numbers when Uncertain} \label{sec:round-answer}

\begin{figure}[ht]
  \centering
  \includegraphics[width=\textwidth]{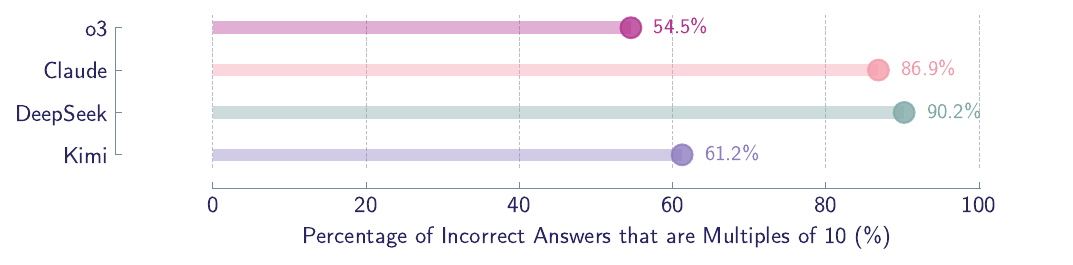}
  \caption{Percentage of incorrect answers that models answer with a number multiple of 10.}
  \label{fig:round-answer}
\end{figure}

In the initial experiments, we observed that models tend to answer round numbers when hallucinating or uncertain. To demonstrate this, in \Cref{fig:round-answer} we show the percentage of incorrect answers that were rounded to the nearest 10. We observe that all models are above 50\%, with some relatively close to 100\%.

\subsection{Separator Comparison} \label{sec:sep-changes}

In the initial experiments, we also ablate whether there are any substantial differences in using different separators between the items of the lists. Thus, we evaluated three different separators: comma (,), semicolon (;), and the pipe (\textbar).

\begin{figure}[h]
  \centering
  \includegraphics[width=\textwidth]{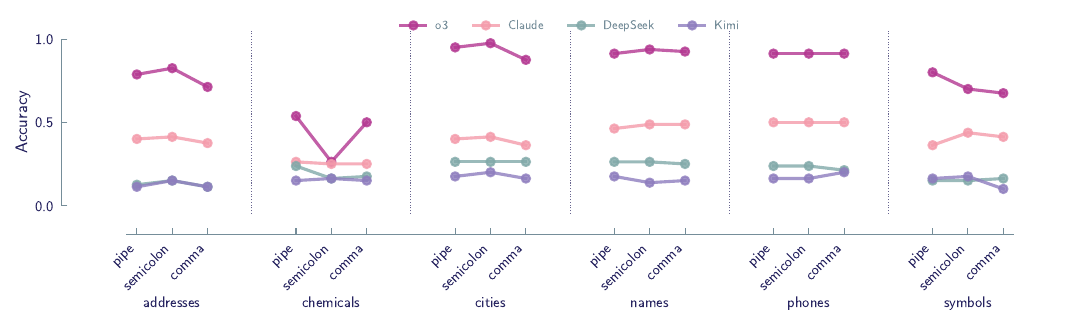}
  \caption{
  Figure showing that the separator used to separate the different semantic classes in the list does not play a big role, and only significant differences are observed for chemicals. For the experiments above, we use the \enquote{pipe} (\textbar).}
  \label{fig:sep-comp}
\end{figure}

\Cref{fig:sep-comp} shows there are no notable differences between separators. As a conclusion of this, we decided to go with the pipe for all the experiments as it is the most intuitive, avoiding possible ambiguities (e.g., with addresses). 

\section{Model Settings} \label{sec:model-settings}

For all of our experiments, we used the recommended settings by the companies that trained each model. \Cref{tab:model-settings} details the configuration for each.

\begin{table}[ht]
\centering
\caption{\textbf{Model settings used in WhatCounts.} The parameters in the table are the only ones modified; everything else we used the default values provided by the providers.}
\begin{tabular}{lcccc}
\hline
\textbf{Parameter} & \textbf{o3} & \textbf{Claude} & \textbf{DeepSeek} & \textbf{Kimi} \\
\hline
Temperature      & 1.0 & 0.0 & 0.6 & 0.6 \\
Top-p            & 1.0 & 1.0 & 0.95 & 1.0 \\
Max tokens       & 100000 & 64000 & 128000 & 16000 \\
Reasoning effort & medium & - & - & - \\
\hline
\end{tabular}
\label{tab:model-settings}
\end{table}

\clearpage

\section{Prompts} \label{sec:prompts}

\begin{lstlisting}[caption=Prompt used for the counting task reported in \Cref{sec:main-results}, label={lst:counting-prompt}]
   {input_text}
   
   How many {field_name()} are above? 
\end{lstlisting}

\noindent where \texttt{input\_text} is the list of items, and \texttt{field\_name} is the specific semantic class to count (e.g., chemical, full person names).

\begin{lstlisting}[caption=Prompt used when mentioning the separator, label={lst:explicit-prompt}]
{input_text}

How many {field_name} separated by '{separator}' are above?"
\end{lstlisting}

\noindent where \texttt{input\_text} is the list of items, \texttt{field\_name} is the specific semantic class to count (e.g., addresses), and \texttt{separator} is the separator used in the list, for our experiments always the \enquote{pipe} (\textbar).

\subsection{Agent Prompts} \label{sec:agent-prompts}

\begin{lstlisting}[caption=System prompt used in the production-ready ablation, label={lst:system-prompt}]
You are an assistant that processes batches of items through an API.

When given a list of items separated by |, you must call submit_batch with:
1. items: The list of items (each item as a separate string, trimmed of whitespace)
2. n_items: The exact count of items

You have access to two tools:
- execute_python: Run Python code to help parse items or count them. Use print() to see output. Python state does not persist.
- submit_batch: Submit the final batch when ready.

The API validates that n_items matches len(items). Count carefully.
You have up to 5 tool calls to complete the task. You MUST call submit_batch to finish.
\end{lstlisting}

\begin{lstlisting}[caption=User prompt in the production-ready ablation, label={lst:user-prompt}]
Process these items by calling submit_batch:

{items_text}
\end{lstlisting}

\noindent where \texttt{items\_text} is the pipe-separated list of items.

\section{Tools Descriptions} \label{sec:tools}

\begin{lstlisting}[caption=Python tool used in the production-ready experiment, label={lst:python-tool}]
{
        "type": "function",
        "function": {
            "name": "execute_python",
            "description": (
                "Execute Python code and return the output. Use this to help parse "
                "items, count them, or verify your work before submitting. "
                "Use print() to see results."
            ),
            "parameters": {
                "type": "object",
                "properties": {
                    "code": {
                        "type": "string",
                        "description": "Python code to execute. Use print() to output results.",
                    }
                },
                "required": ["code"],
            },
        },
    }
\end{lstlisting}

\begin{lstlisting}[caption=Submit batch tool used in the production-ready experiment, label={lst:submit-tool}]
    {
        "type": "function",
        "function": {
            "name": "submit_batch",
            "description": (
                "Submit a batch of items for processing. You MUST provide both the items "
                "and the exact count of items. The count is used as an integrity check. "
                "Call this when you are ready to submit your final answer."
            ),
            "parameters": {
                "type": "object",
                "properties": {
                    "items": {
                        "type": "array",
                        "items": {"type": "string"},
                        "description": "The list of items to process",
                    },
                    "n_items": {
                        "type": "integer",
                        "description": "The number of items in the list (must match len(items) exactly)",
                    },
                },
                "required": ["items", "n_items"],
            },
        },
    }
\end{lstlisting}

\clearpage
\section{Detailed Results}

\subsection{Overall Results}

\begin{figure}[ht]
  \centering
  \includegraphics[width=\textwidth]{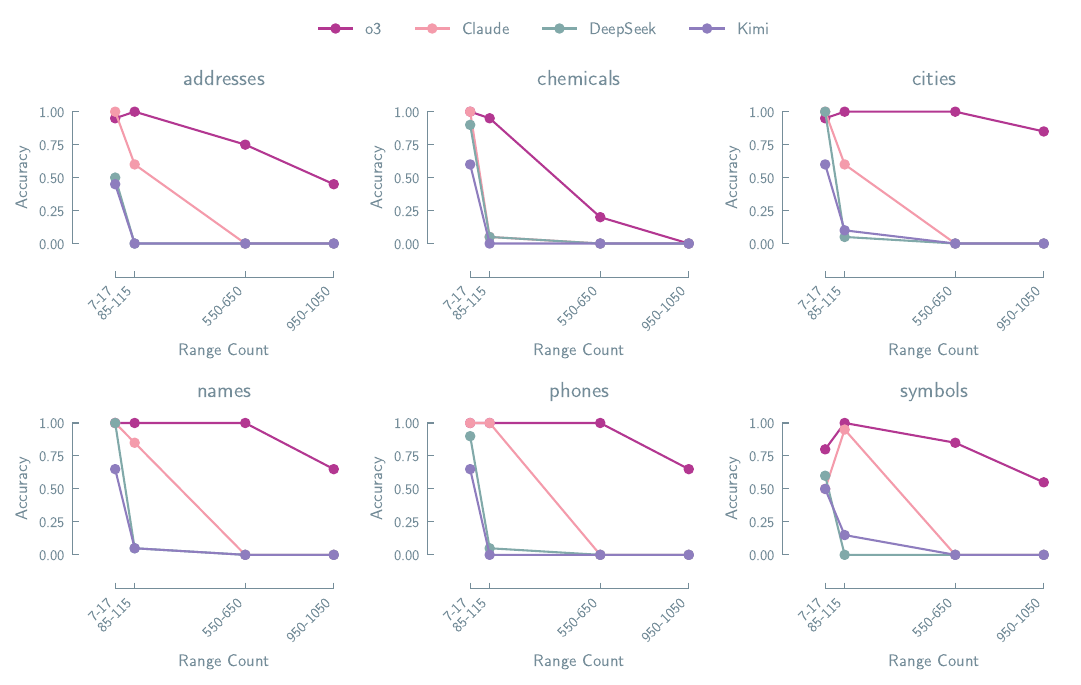}
  \caption{\textbf{Detailed results for the counting task.} The results show very different behaviors for the different semantic classes that are consistent over the evaluated models.
  }
  \label{fig:general_app}
\end{figure}

\Cref{fig:general_app} shows the detailed results for the counting task for each of the six evaluated semantic classes, and over the four ranges that we considered.

We observe that for the semantic classes with the highest semantic gaps, the accuracy scores fall already for shorter ranges, showing even not really good scores for the shortest range (counts between seven and seventeen items).

\clearpage
\subsection{Token-count Dependence}

We run the ablation by selecting items that fit within a range of tokens rather than just picking the count. The token ranges considered were (99, 101), (999, 1001), and (9999, 10001), giving a 10~\% margin. This is that, for example, when randomly picking items for the cities list, it will sample until the list token count falls within the range 99 to 101 tokens, plus or minus the 10~\% range. The margin was added to make it easier for the algorithm to find lists within the range, especially for token-heavy classes such as addresses. As in the previous experiment, and in all experiments we run, for each range and semantic class, 20 different questions are used to account for statistical factors. 

\begin{figure}[h]
  \centering
  \includegraphics[width=\textwidth]{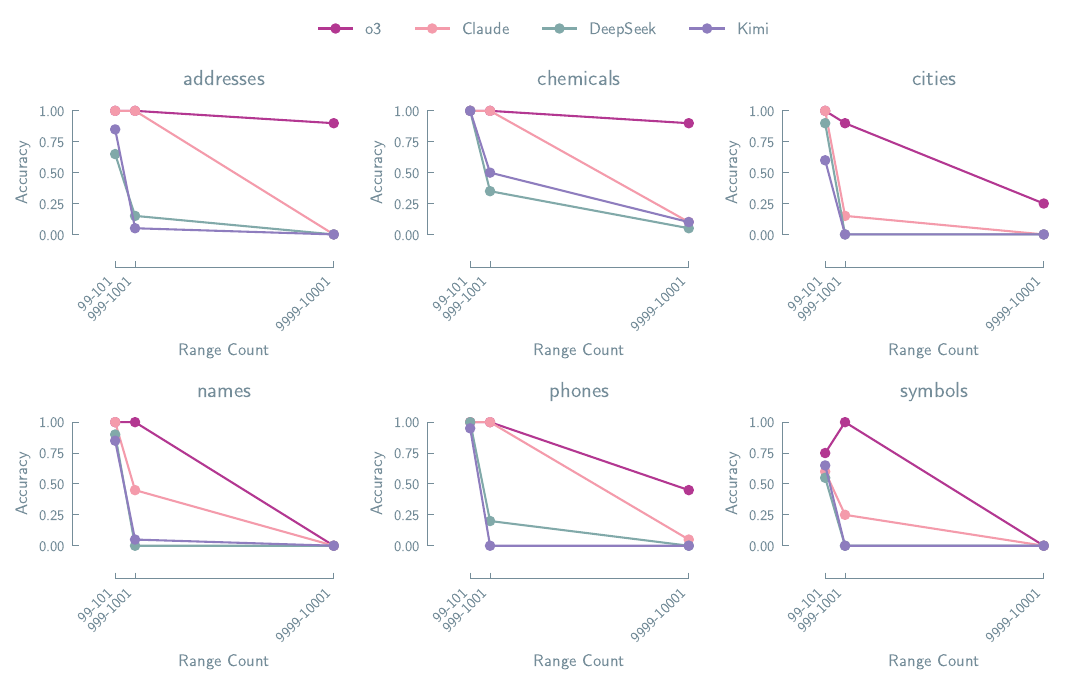}
  \caption{\textbf{Figure representing the token-count control ablation experiment.} We observe that despite the trends being different compared to the original counting task, invariance between the different semantic classes still exists for all the models.
  }
  \label{fig:token_app}
\end{figure}

\Cref{fig:token_app} shows the results of fixing the number of tokens in the list. We see that there is still significant invariance across classes, demonstrating that the token count cannot explain the semantic fragility we report.

\clearpage
\subsection{Explicit Separator Dependence}

To ensure that the identification of the separator used in the list poses no problem to the models, we run the ablation experiment in which we modify the input prompt to explicitly mention the separator that is used in the list.

\begin{figure}[h]
  \centering
  \includegraphics[width=\textwidth]{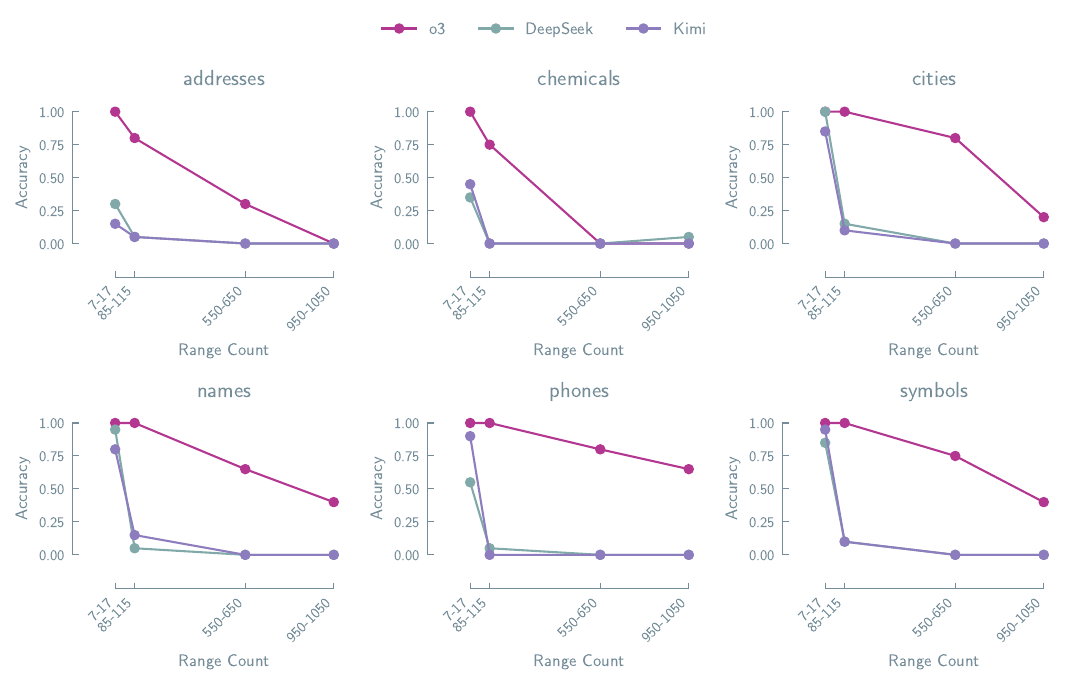}
  \caption{\textbf{Results for the ablation of explicitly mentioning the separator in the prompt.} We observe no differences with respect to the counting task when the separator is not mentioned.}
  \label{fig:explicit_app}
\end{figure}

\Cref{fig:explicit_app} shows the extended results when the models are prompted to count while also being told what the separator used was. The results show still a very high variance among the different variables, resulting in very similar trends as when not mentioning the separator.

\clearpage
\subsection{Identification Experiment}

To evaluate how good the identification of the models is, and thus clarify if models fail because of this or because of the aggregation, we designed the ablation of prompting the models to return the list with the items wrapped between XML tags.

\begin{figure}[h]
  \centering
  \includegraphics[width=\textwidth]{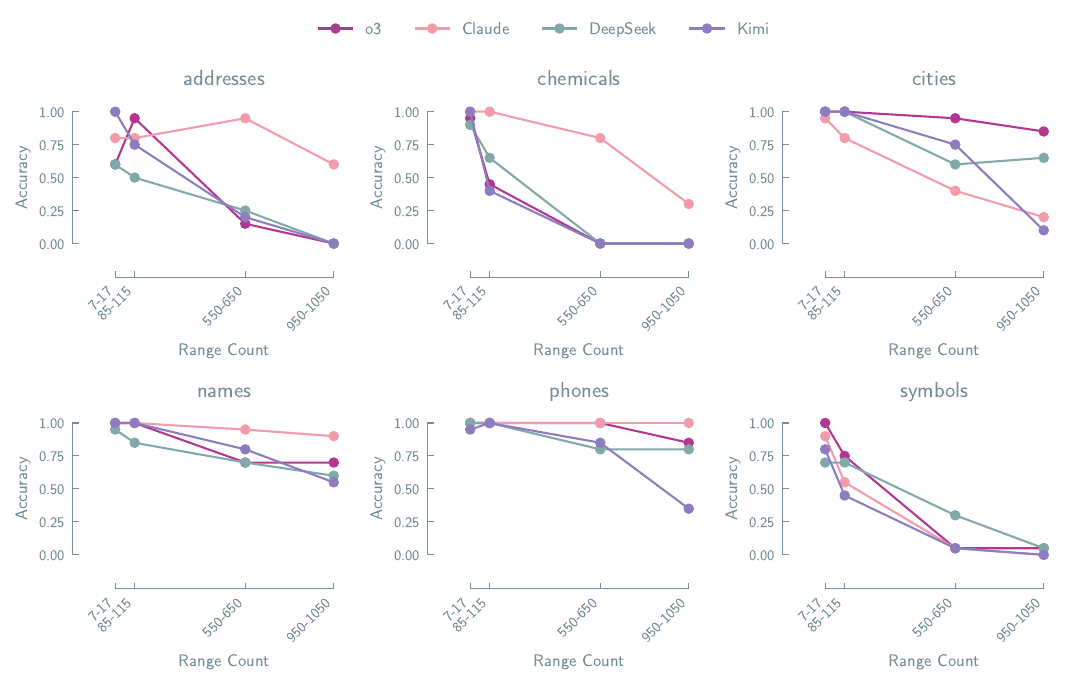}
  \caption{\textbf{Figure representing the wrapping the items ablation}. The scores are detailed for each class and each of the four ranges. We observe that the models are able to identify the items better than they can count them (\Cref{fig:general_app}).}
  \label{fig:wrap_app}
\end{figure}

\Cref{fig:wrap_app} shows the results for the wrapping ablation, detailed for each semantic class and for the ranges used. These results show that the models are better at identifying items in the list than at counting them. However, there is still a significant semantic gap, as the performance on this task for these ablations varies across semantic classes.

\clearpage
\subsection{Aggregation Detailed Results}

To evaluate the aggregation ability of the models when doing the counting task, and trying to identify the underlying reason for the semantic variability, we wrapped the items in the input lists between XML tags.

\begin{figure}[h]
    \centering
    \includegraphics[width=\textwidth]{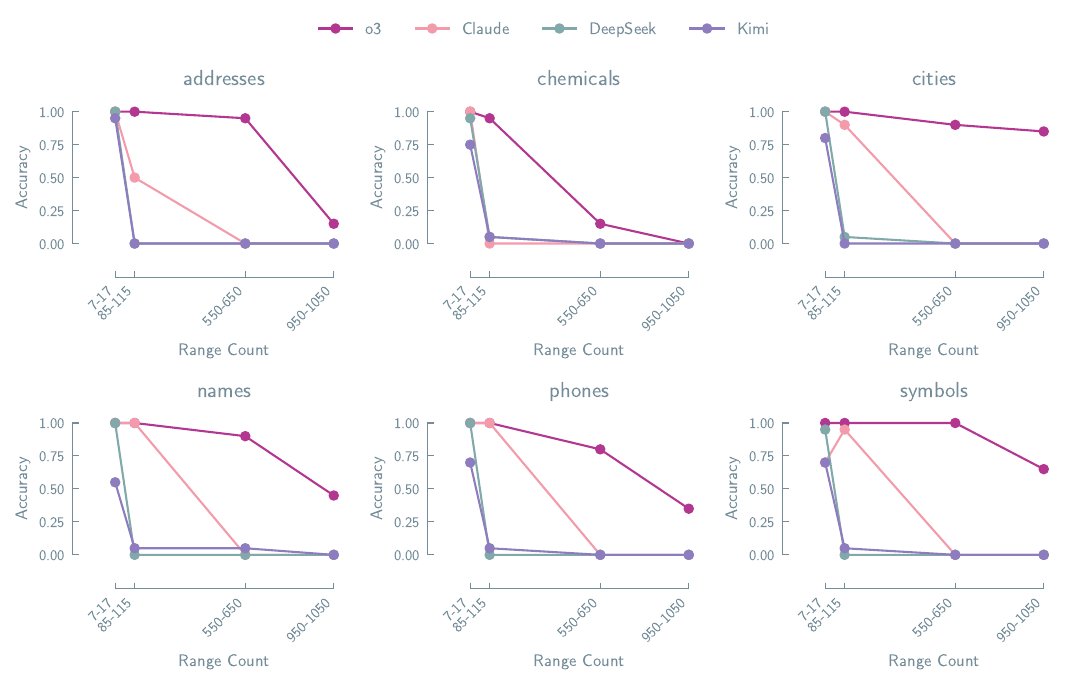}
    \caption{\textbf{Aggregation of clearly identified items still show sustantial semantic dependence.} We observe no big differences in the trends when prompted the models to count without XML tags wrapping the items.}
    \label{fig:xml-app}
\end{figure}

\clearpage
\subsection{Token-distribution Dependence} \label{sec:shuffling-app}

To determine how the token distribution in each semantic class affects the performance in the counting task, we run the experiment in which we shuffle the tokens of each item in the list.

\begin{figure}[h]
  \centering
  \includegraphics[width=\textwidth]{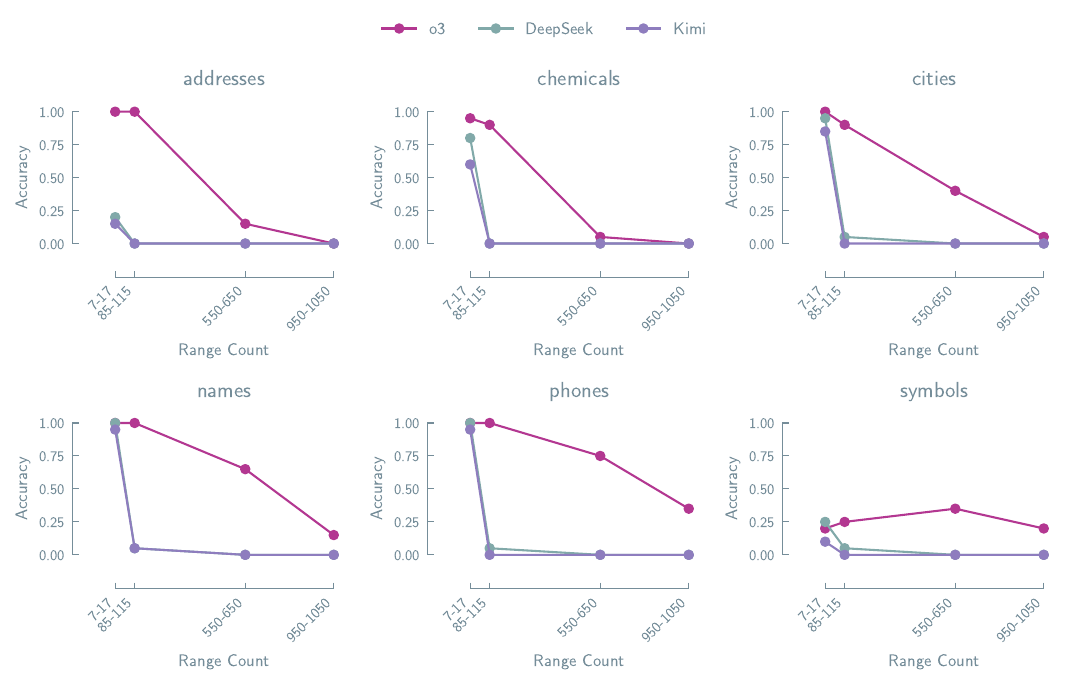}
  \caption{\textbf{Detailed results for the ablation when shuffling the tokens of the items in the list.} We see that there is still a big variance between the different semantic classes, suggesting that the token distribution does not affect all classes the same way.
  }
  \label{fig:shuffle_app}
\end{figure}

\Cref{fig:shuffle_app} shows the results when shuffling the tokens of the items in the list. It is possible to see that the performance still changes for the different semantic classes.

\begin{figure}[h]
  \centering
  \includegraphics[width=\textwidth]{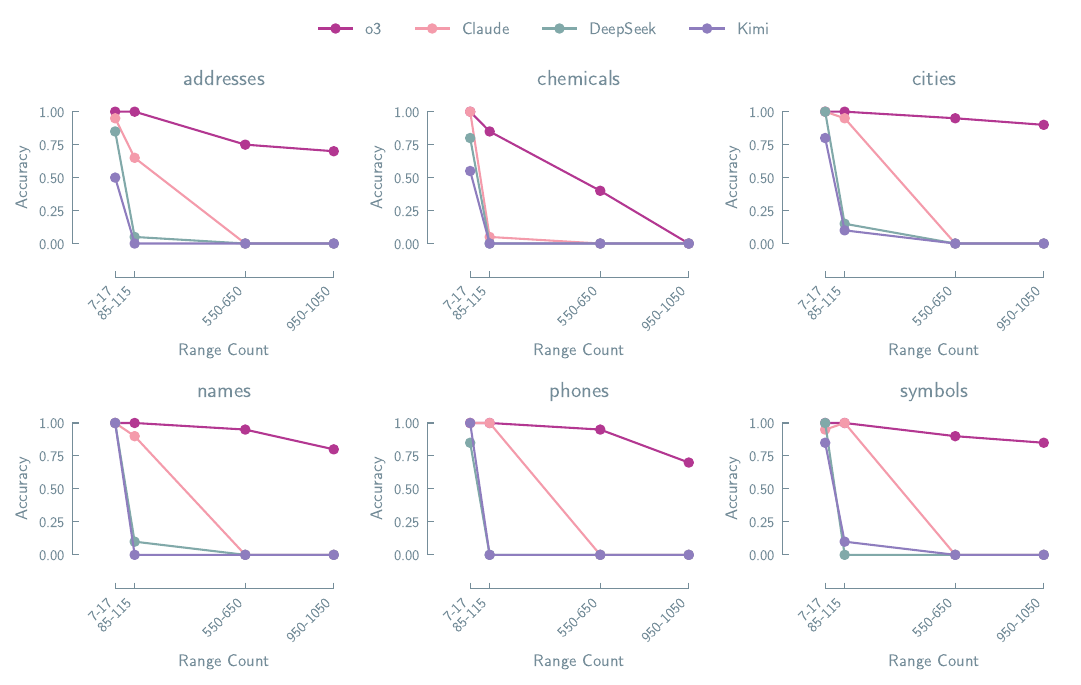}
  \caption{\textbf{Detailed results for the ablation when shuffling the tokens of the items in the list, and explicitly mentioning the separator in the prompt.}}
  \label{fig:shuffle_ent_app}
\end{figure}

As shuffling without mentioning the separator might seem unfair, since the items might have lost all meaning. Thus, we run the same experiment but using the prompt in \Cref{lst:explicit-prompt}. The results in \Cref{fig:shuffle_ent_app} show no big differences with respect to the counting task without shuffling, suggesting that models learn the underlying token distribution rather than the meaning of those tokens.

\clearpage

\section{Semantic class Complexity}

\begin{table}
    \centering
    \caption{\textbf{Table showing the complexity distribution for each of the studied semantic classes.} The table illustrates the complexity of each of the semantic classes using two metrics: the Kolmogorov complexity, which measures how compressible the classes (per entity) or list of items (full input) are, and the number of tokens per class. Chemicals are more complex according to both metrics because of the single-character tokens present in the IUPAC notation, such as brackets, numbers, and other symbols.}
    \begin{tabular}{lccc}
\toprule
\multirow{2}{*}{Entity Type} &
\multicolumn{2}{c}{Kolmogorov complexity} &
\multirow{2}{*}{Tokens/Entity} \\
\cmidrule(lr){2-3}
& Full input & Per entity & \\
\midrule
Addresses & 10818.52 & 59.65 & 13.99 \\
Chemicals & 9644.51 & 89.90 & 41.15 \\
Cities & 3053.21 & 29.89 & 3.32 \\
Names & 3286.52 & 35.48 & 4.76 \\
Phones & 3571.99 & 33.85 & 7.38 \\
Symbols & 1888.51 & 24.12 & 2.85 \\
\midrule
Overall & 5377.21 & 45.84 & 12.33 \\
\bottomrule
\end{tabular}

    \label{tab:token-complexity}
\end{table}

To study whether the token counts for the different classes affect the count, we can check them across all the lists for which the models were evaluated (excluding those from Claude, as its tokenizer is not available). Thus, in \Cref{tab:token-complexity} we show the tokens per entity for the different classes, and in addition, the Kolmogorov complexity, that have been suggested as a measure of what LLMs \enquote{prefer} \cite{goldblum2023free}.

While the complexity of an item (like a long chemical formula) does make tasks harder, it does not fully explain our results. If complexity were the only factor, performance would steadily decline as items become more complex. However, we find that simple items like symbols can be surprisingly difficult, while some complex items like phone numbers are managed well. 

\clearpage
\section{DPO and PPO Comparison}

The work of \citet{ivison2024unpacking} trained six different models in 60K data regimes using the same training configuration and only changing the dataset and training method, three of them using DPO, and three using PPO. \Cref{tab:dpo-ppo-comparison} details the accuracy observed for the six of them, and the difference between them. We observe no big differences between different methods for the same datasets-based model.

\begin{table}[h]
    \centering
    \caption{\textbf{Comparison between the evaluation of DPO and PPO models}. The table shows the accuracy of the DPO and PPO models, as well as the difference between those trained on the same dataset. The results are averaged across all ranges and detailed for each class.}
    \begin{tabular}{lccccccc}
\toprule
Model & Addresses & Chemicals & Cities & Names & Phones & Symbols & Overall \\
\midrule
Base & 0.16 & 0.21 & 0.24 & 0.31 & 0.20 & 0.19 & 0.22 \\
\midrule
\textbf{StackExchange} &  &  &  &  &  &  &  \\
DPO & 0.12 & 0.19 & 0.17 & 0.30 & 0.21 & 0.20 & 0.20 \\
PPO & 0.10 & 0.21 & 0.19 & 0.28 & 0.20 & 0.20 & 0.20 \\
$\Delta$ (DPO - PPO) & +0.02 & -0.02 & -0.01 & +0.02 & +0.01 & 0.00 & +0.00 \\
\midrule
\textbf{HH-RLHF} &  &  &  &  &  &  &  \\
DPO & 0.09 & 0.20 & 0.19 & 0.28 & 0.30 & 0.19 & 0.21 \\
PPO & 0.11 & 0.23 & 0.16 & 0.30 & 0.23 & 0.16 & 0.20 \\
$\Delta$ (DPO - PPO) & -0.02 & -0.03 & +0.02 & -0.02 & +0.07 & +0.02 & +0.01 \\
\midrule
\textbf{Nectar} &  &  &  &  &  &  &  \\
DPO & 0.09 & 0.24 & 0.21 & 0.24 & 0.25 & 0.19 & 0.20 \\
PPO & 0.10 & 0.20 & 0.25 & 0.24 & 0.29 & 0.18 & 0.21 \\
$\Delta$ (DPO - PPO) & -0.01 & +0.04 & -0.04 & 0.00 & -0.04 & +0.01 & -0.01 \\
\bottomrule
\end{tabular}

    \label{tab:dpo-ppo-comparison}
\end{table}

\section{Finetunings Performance in GSM8K} \label{sec:gsm8k-runs}

To evaluate how the same training regime, but on different datasets behave in other popular benchmarks, we run the same models from \citet{ivison2024unpacking} in GSM8K \cite{cobbe2021training}. 

\begin{table}[h]
    \centering
    \caption{\textbf{Results of running the finetuned models in GSM8K.} We observe that the differences between models are not very big, as the semantic gap variation in WhatCounts.}
    \begin{tabular}{lcc}
\toprule
Model & Accuracy (\%) & Difference (\%) \\
\midrule
Base & 46.17 & -- \\
\midrule
DPO-HH & 45.03 & -1.14 \\
DPO-Nectar & 43.75 & -2.43 \\
DPO-Stack & 44.88 & -1.29 \\
PPO-HH & 47.08 & +0.91 \\
PPO-Nectar & 46.10 & -0.08 \\
PPO-Stack & 45.87 & -0.30 \\
\bottomrule
\end{tabular}
    \label{tab:gsm8k-runs}
\end{table}

\clearpage
\section{Location and Gender Analysis} \label{sec:bias-analysis}

To study whether the different origins of some of the semantic classes might have an influence on the accuracy of the models, for the classes of names, addresses, cities, and phones, we created tasks created solely on different locations (the US, Great Britain, Spain, France, Germany, Japan, and China). This, for example, limits the list to only those from the US, or for the other locations.

\begin{figure}
    \centering
    \includegraphics[width=\textwidth]{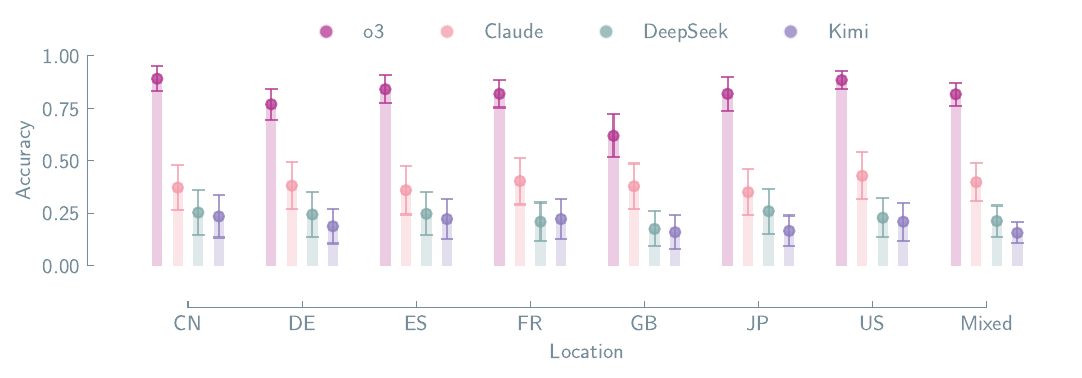}
    \caption{\textbf{Results of evaluating on specific locations.} Mixed stands for the experiment reported in \Cref{fig:fig-1} where the items were picked by sampling from all the locations. The scores in the plot are the average across the four evaluated classes in this experiment (addresses, cities, names, and phones), along with the different ranges. The results show that there are no trends among the evaluated models, indicating that classes based on a specific country do not lead to a larger semantic gap.}
    \label{fig:location-based}
\end{figure}

\Cref{fig:location-based} shows that when focusing the evaluation only on the semantic classes based on a simple location or country, no trends are observed, suggesting that different location-based classes do not increase the semantic gap.

Similarly, we also investigated whether gender can have an influence on the semantic gap observed for the named semantic class. For this, the lists of this experiment are specifically created using names from one gender or the other gender.

\begin{figure}
    \centering
    \includegraphics[width=\textwidth]{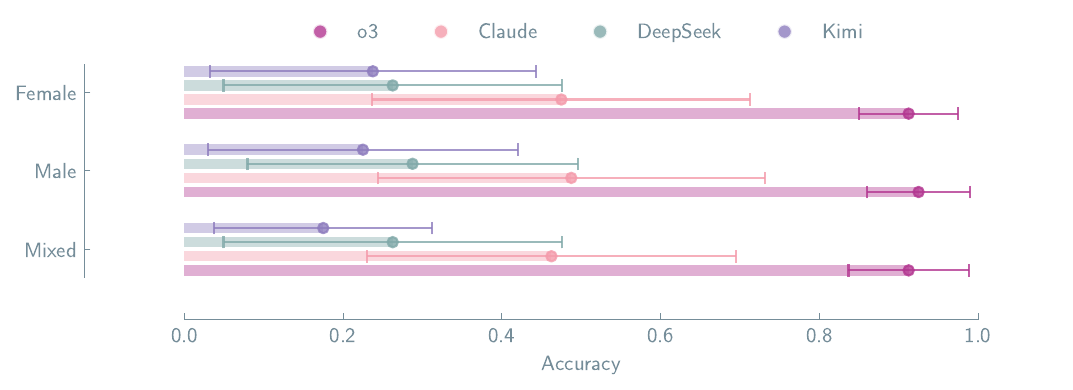}
    \caption{\textbf{Results of evaluating on one specific gender}. The figure shows the results for each gender, and when mixed (\Cref{fig:fig-1}). Note that this experiment only considers the class names. We observe no trends among models, suggesting that gender does not contribute to or influence the semantic gap in any way.}
    \label{fig:gender-based}
\end{figure}

\Cref{fig:gender-based} on evaluating the effect when only considering one of the genders when evaluating counting names. It shows that there is no trend between the models, which means that the gender of the names does not influence the semantic gap.

\clearpage

\end{document}